\documentclass{article}

\PassOptionsToPackage{numbers, compress}{natbib}

\usepackage[final]{neurips_2022}

\usepackage[utf8]{inputenc} 
\usepackage[T1]{fontenc}    
\usepackage{hyperref}       
\usepackage{url}            
\usepackage{booktabs}       
\usepackage{amsfonts}       
\usepackage{nicefrac}       
\usepackage{microtype}      
\usepackage{xcolor}         

\usepackage{times}
\usepackage{epsfig}
\usepackage{graphicx}
\usepackage{amsmath}
\usepackage{amssymb}
\usepackage{algorithm}
\usepackage{algpseudocode}
\usepackage{threeparttable}
\usepackage{multirow}
\usepackage{booktabs}
\usepackage{caption}
\usepackage{mathrsfs}
\usepackage{bbding}
\usepackage{enumitem}
\usepackage{arydshln}

\title{Fine-Grained Semantically Aligned \\ Vision-Language Pre-Training}

\author{Juncheng Li$~\textsuperscript{\rm 1, 2}$\thanks{Work done when interning at Huawei Cloud.} \ \thanks{Equal Contribution.}  \And Xin He$~\textsuperscript{\rm 2}$\footnote[2]{} \And Longhui Wei$~\textsuperscript{\rm 2}$\footnote[2]{}  \And  Long Qian$~\textsuperscript{\rm 1}$ \And  Linchao Zhu$~\textsuperscript{\rm 1}$ \vspace{-1cm} \AND \vspace{-3cm}   Lingxi Xie$~\textsuperscript{\rm 2}$ \And Yueting Zhuang$~\textsuperscript{\rm 1}$\thanks{Corresponding Authors.}  \And Qi Tian$~\textsuperscript{\rm 2}$\footnote[3]{}   \And  Siliang Tang$~\textsuperscript{\rm 1}$\footnote[3]{}  \And \vspace{-0.5cm}  \\
	\small{$~\textsuperscript{\rm 1}$ Zhejiang University}, 
	\small{$~\textsuperscript{\rm 2}$ Huawei Cloud}
	\\{\tt\small {\{junchengli, qianlong0926, yzhuang, siliang\}}@zju.edu.cn}
	\\{\tt\small {\{hexin80, weilonghui1, tian.qi1\}}@huawei.com}
	\\{\tt\small {\{zhulinchao7, 198808xc\}}@gmail.com}
}

\begin{document}
	\normalsize
	
	\maketitle

	\begin{abstract}
		Large-scale vision-language pre-training has shown impressive advances in a wide range of downstream tasks. Existing methods mainly model the cross-modal alignment by the similarity of the global representations of images and texts, or advanced cross-modal attention upon image and text features. However, they fail to explicitly learn the fine-grained semantic alignment between visual regions and textual phrases, as only global image-text alignment information is available. In this paper, we introduce \textbf{\textsl{LOUPE}}\includegraphics[height=0.106in, width=0.11in]{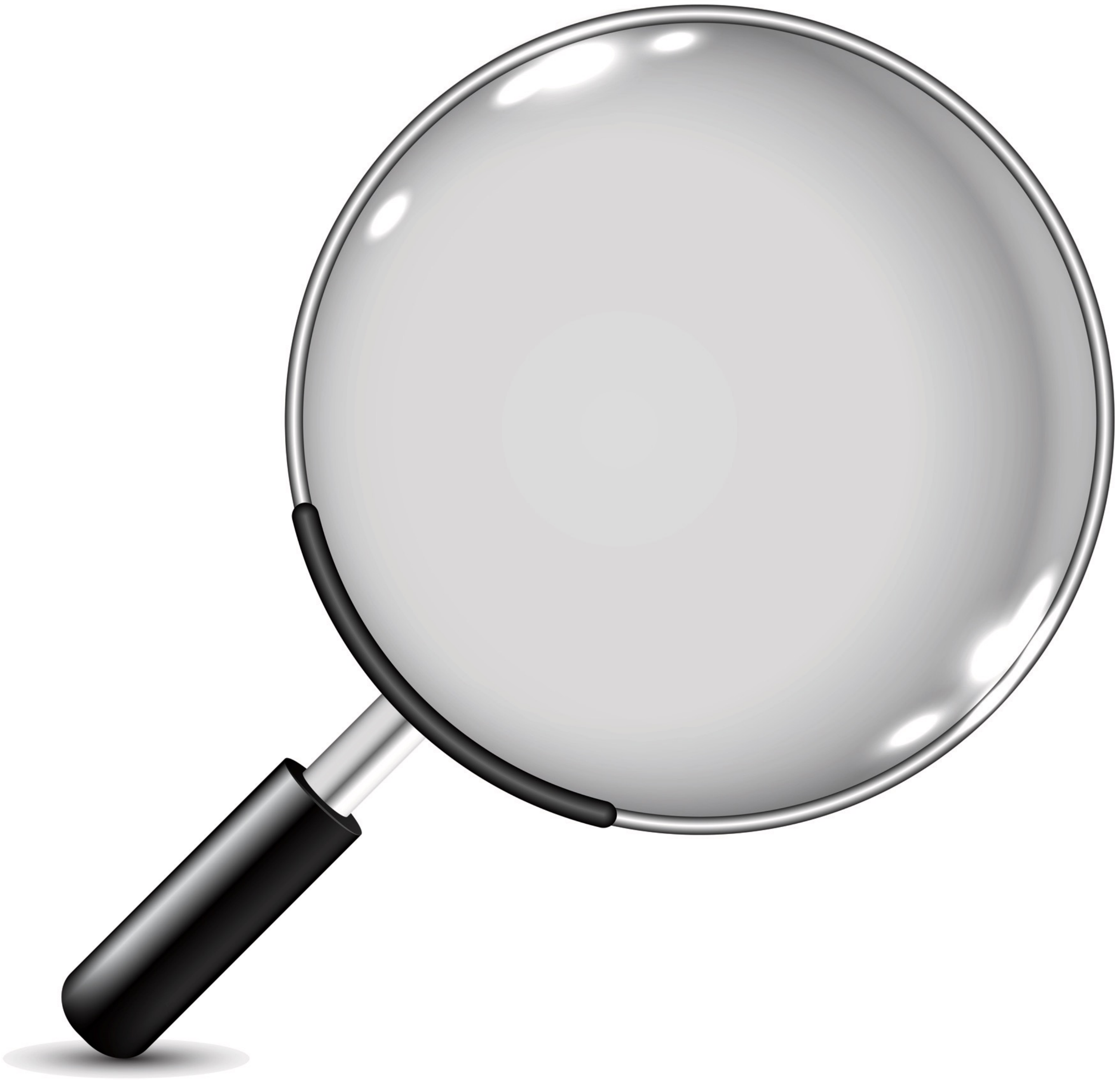}, a fine-grained semantically a\textbf{L}igned visi\textbf{O}n-lang\textbf{U}age \textbf{P}r\textbf{E}-training framework, which learns fine-grained semantic alignment from the novel perspective of game-theoretic interactions. To efficiently compute the game-theoretic interactions, we further propose an uncertainty-aware neural Shapley interaction learning module. Experiments show that LOUPE achieves state-of-the-art performance on a variety of  vision-language tasks. Furthermore, without any object-level human annotations and fine-tuning, LOUPE achieves competitive performance on object detection and visual grounding. More importantly, LOUPE opens a new promising direction of learning fine-grained semantics from large-scale raw image-text pairs. The repository of this work is at \url{https://github.com/YYJMJC/LOUPE}.
	\end{abstract}

	\section{Introduction}
	
	Learning transferable cross-modal representations from large-scale vision-language pre-training has exhibited remarkable performance on a wide variety of downstream tasks. Most existing works can be classified into two categories:  \emph{dual-encoder} and \emph{fusion-encoder}. The dual-encoder methods~\cite{jia2021scaling, li2021supervision, radford2021learning, yao2021filip} adopt two separate encoders to embed images and texts, and model the cross-modal alignment by the cosine similarity between the global features of images and texts. While such architecture is efficient for large-scale image-text retrieval by pre-computing image and text representations offline, they fail to model fine-grained semantic alignment between visual regions and textual phrases. On the other hand, the fusion-encoder methods~\cite{chen2020uniter, kim2021vilt, li2021align, li2020oscar, lu2019vilbert, qi2020imagebert, tan2019lxmert, su2019vl} attempt to use a single multi-modal encoder to jointly model the concatenated sequence of images and texts. These methods simulate soft alignment via advanced cross-modal attention~\cite{vaswani2017attention}. However, they can only learn implicit alignment by end-to-end training, lacking explicit supervision to encourage semantic alignment between visual regions and textual phrases. And the learned cross-modal attention matrices are often scattering and uninterpretable. Further, they are inefficient for retrieval since it requires jointly encoding every image-text pair during inference.
	
	Learning fine-grained semantic alignment from image-text pre-training is crucial to many cross-modal reasoning tasks (\textsl{e.g., visual grounding}~\cite{yu2016modeling},\textsl{ image captioning}~\cite{xu2015show}), but it is particularly challenging as the alignment information between visual regions and textual phrases is not available, posing fine-grained semantic alignment learning a weakly-supervised learning problem. In this paper, we address this problem while simultaneously maintaining high retrieval efficiency by proposing \textbf{\textsl{LOUPE}}\includegraphics[height=0.106in, width=0.11in]{loupe.pdf}, a fine-grained semantically a\textbf{L}igned visi\textbf{O}n-lang\textbf{U}age \textbf{P}r\textbf{E}-training framework, from the novel perspective of game theory. We formulate input patch and word tokens as multiple players into a cooperative game and quantify game-theoretic interactions~(\textsl{i.e., Shapley Interaction} ~\cite{grabisch1999axiomatic, shapley1953value}) among them to investigate the semantic alignment information. LOUPE learns fine-grained semantic alignment from two stages:  \emph{token-level Shapley interaction modeling} and \emph{semantics-level Shapley interaction modeling}, where we first learn to identify semantic regions of images that correspond to some semantically meaningful entities, and then align these regions with phrases in the paired text.

	Specifically, \emph{token-level Shapley interaction modeling} aims to group patch tokens of images into semantic regions that semantically correspond to some visual instances. From the game-theoretic view, we take patch tokens as players and the similarity score between images and texts as the game function. 
	Intuitively, supposing a set of patch tokens correspond to a visual instance in the image, then they tend to have strong interaction to form the complete semantics of the corresponding instance, which contributes to the better similarity judgment with the paired text. 
	Based on this insight, we take the token-level Shapley interaction as soft supervision labels to encourage the model to capture semantic regions from images. Then, \emph{semantics-level Shapley interaction modeling} infers the fine-grained semantic alignment between semantic regions and phrases. We consider every region and phrase as players and define a fine-grained similarity score as the game function. If a region and a phrase have strong correspondence, they tend to interact with each other and contribute to the fine-grained similarity score. By measuring the Shapley interaction between each region-phrase pair, we obtain the alignment information to guide the pre-training model. 
	
	As computing the exact Shapley interaction is an NP-hard problem~\cite{matsui2001np}, existing methods mainly employ sampling-based method~\cite{castro2009polynomial} to obtain unbiased estimation. However, as the number of players grows, they require thousands of model evaluations. To reduce the computational cost, we further propose an efficient hybrid Shapley interaction learning strategy, where an uncertainty-aware neural Shapley interaction learning module cooperates with the sampling-based method. Experimental results show that our hybrid strategy significantly saves the computational cost while maintaining the estimation accuracy. More analysis is shown in Section~\ref{4.4}.

	Our framework serves as a proxy training objective that explicitly establishes the fine-grained semantic alignment between local region and phrase representations. This proxy objective can be directly removed for downstream tasks, rendering an efficient and semantics-sensitive \emph{dual-encoder} model. Experiments show that LOUPE achieves new state-of-the-art on image-text retrieval benchmarks. For text-to-image retrieval on MSCOCO, LOUPE surpasses its strongest competitor by 4.2\% on recall@1. Further, without any fine-tuning, LOUPE successfully transfers to object detection and visual grounding tasks in a zero-shot manner. For object detection, it achieves 12.1\% mAP on COCO and 19.5\% mAP on PASCAL VOC. For visual grounding, it achieves 26.8\% accuracy on RefCOCO and 23.6\% accuracy on RefCOCO+. Our contributions are summarized as follows:
	
	\begin{itemize}[leftmargin=0.8cm]
		
		\item We propose \textbf{\textsl{LOUPE}}\includegraphics[height=0.106in, width=0.11in]{loupe.pdf} that explicitly learns fine-grained semantic alignment between visual regions and textual phrases while preserving the high retrieval efficiency of dual-encoder.
		
		\item We introduce an efficient and effective hybrid Shapley interaction learning strategy, based on an uncertainty-aware neural Shapley interaction learning module and a sampling-based method.
		
		\item Pre-trained on image-text data, LOUPE achieves new state-of-the-art on image-text retrieval and successfully transfers to the tasks that require more fine-grained object-level visual understanding (\textsl{i.e., object detection and visual grounding}) without any fine-tuning.
		
		\item As manual annotations for masses of object categories is time-consuming and unscalable, our work demonstrates a promising alternative, that is, learning fine-grained semantics from raw texts about images, which are easily available and contain a broader set of visual concepts.
		
	\end{itemize}

	\section{Related Work}
	\textbf{Vision-Language Pre-Training.} The great success of  pre-train-and-fine-tune paradigm in natural language processing~\cite{brown2020language, devlin2018bert} and computer vision~\cite{dosovitskiy2020image, he2020momentum, wei2022mvp} has been expanded to the joint domain of vision and language~\cite{anderson2018bottom, antol2015vqa, li2020unsupervised}. Dominant vision-language pre-training models can be categorized into two groups: \emph{dual-encoder} and \emph{fusion-encoder}. The dual-encoder methods~\cite{jia2021scaling, li2021supervision, radford2021learning, yao2021filip} adopt two individual encoders to embed images and texts separately, and model the cross-modal interaction by cosine similarity. Such architecture is efficient for large-scale image-text retrieval as image and text representations can be pre-computed offline. However, simply measuring the cosine similarity between global representations is shallow to capture fine-grained semantic relationships between regions and phrases. 
	The fusion-encoder methods~\cite{chen2020uniter, huang2021seeing, huang2020pixel, kim2021vilt, li2021align, li2020unimo, li2020oscar, lu2019vilbert, qi2020imagebert, su2019vl, tan2019lxmert, yu2020ernie, zhang2021vinvl} adopt a single multi-modal encoder to jointly model the concatenated sequence of images and texts, which achieves deeper cross-modality interaction. However, these methods are less efficient as images and texts are intertwined to compute the cross-modal attention and can not be pre-computed offline. Further, there are no explicit supervision signals to encourage the alignment between regions and phrases. Some works~\cite{chen2020uniter, li2020unimo, li2020oscar, lu2019vilbert, tan2019lxmert, yu2020ernie, zhang2021vinvl, zhong2022regionclip} attempt to leverage an off-the-shelf object detector  to extract object features for pre-training. However, the detector is usually pre-trained on limited object categories.
	Furthermore, considering the excessive demand on memory and computation, existing methods usually fix the parameters of detection models and regard region detection as a pre-processing step, disconnected with vision-language pre-training. Thus, the performance is also restricted by the quality of  detection models.
	FILIP~\cite{yao2021filip} uses a token-wise maximum similarity to enhance the cross-modal interaction of dual-encoder methods. 
	To learn explicit fine-grained semantic alignment, GLIP~\cite{li2021grounded} and X-VLM~\cite{zeng2021multi} utilize human-annotated datasets, where regions with bounding-box annotations are aligned with text descriptions. Such a manner is time-consuming and hard to scale to larger raw image-text data from the Internet.
	In contrast, our proposed framework explicitly learns the fine-grained semantic alignment from raw image-text data and at the same time maintains the high efficiency of dual-encoder. Detailed discussions can be found in Appendix K.

	\textbf{Shapley Values.} The Shapley value~\cite{shapley1953value} was initially introduced in game theory. It has been theoretically proven to be the unique metric to fairly estimate the contribution of each player in a cooperative game such that certain desirable axioms are satisfied~\cite{weber1988probabilistic}. With solid theoretic foundations, Shapley value has recently been studied as post-hoc explanation methods for Deep Neural Networks~(DNN)~\cite{datta2016algorithmic, lundberg2017unified, zhang2020interpreting}. Lundberg \textsl{et al.}~\cite{lundberg2017unified} propose a unified attribution method based on Shapley value to interpret the predictions of DNN. Ren~\textsl{et al.}~\cite{ren2021unified} propose to explain adversarial attacks by Shapley value. In this paper, we propose to model fine-grained semantic alignment by game-theoretic interactions, along with an efficient Shapley interaction learning strategy.

	\section{Method}
	In this section, we first introduce the problem formulation of fine-grained semantically aligned vision-language pre-training in Section \ref{3.1}. Then, we propose the corresponding LOUPE framework for fine-grained semantic alignment learning in Section \ref{3.2} and an efficient approach for Shapley interaction learning in Section \ref{3.3}. 
	
	\subsection{Problem Formulation and Model Overview}
	\label{3.1}
	
	Generally, vision-language pre-training aims to learn an image encoder $f_\mathrm{I}$ and a text encoder $f_\mathrm{T}$ by cross-modal contrastive learning, where the matched image-text pairs are optimized to get closer and the mismatched pairs are optimized to get further. Let $f_\mathrm{I}(I_i)$ and $f_\mathrm{T}(T_i)$ denote the global representations of the image and text. Then the cross-modal contrastive loss can be formulated as:
	
	\begin{equation}
		\mathcal{L}_\mathrm{CMC} = - \log\frac{\exp(f_\mathrm{I}(I_i)^\top f_\mathrm{T}(T_i)/\tau))}{\sum_{j}^B \exp(f_\mathrm{I}(I_i)^\top f_\mathrm{T}(T_j)/\tau)} -  \log\frac{\exp(f_\mathrm{I}(I_i)^\top f_\mathrm{T}(T_i)/\tau))}{\sum_{j}^B \exp(f_\mathrm{I}(I_j)^\top f_\mathrm{T}(T_i)/\tau))} 
	\end{equation}
	where $B$ is the batch size and $\tau$ is the temperature hyper-parameter.
	
	While intuitive, such a manner can only learn coarse alignment between images and texts but fails to explicitly capture the fine-grained semantic alignment between visual regions and textual phrases. To learn fine-grained semantic alignment  while simultaneously maintaining high retrieval efficiency, we propose LOUPE, a fine-grained semantically aligned vision-language pre-training framework that germinates from cooperative game theory. 
	
\begin{figure}[!t]
		\centering
		\includegraphics[width=\textwidth]{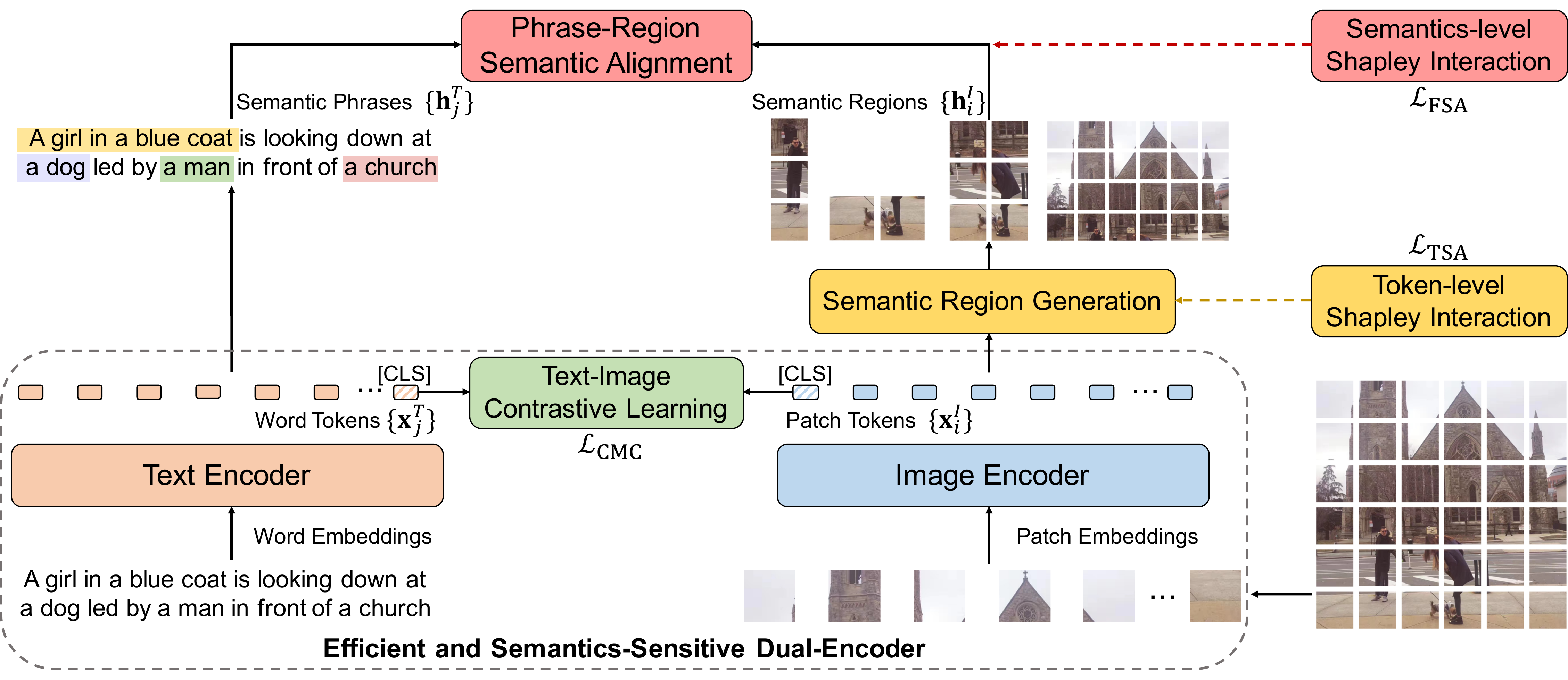}
		\vspace{-0.4cm}
		\caption{Overview of LOUPE. Our framework serves as a proxy training objective that encourages the image encoder to capture semantic regions and establishes the semantic alignment between region and phrase representations. The proxy training objective can be easily removed for downstream tasks, rendering an efficient and semantics-sensitive dual-encoder.}
		\label{framework}
		\vspace{-0.5cm}
	\end{figure}

	As illustrated in Figure \ref{framework}, LOUPE learns fine-grained semantic alignment from two stages: \textbf{token-level Shapley interaction modeling} and \textbf{semantics-level Shapley interaction modeling}. For token-level Shapley interaction modeling, we learn to aggregate patch tokens of images into semantic regions that semantically correspond to some visual concepts, under the guidance of token-based semantic aggregation loss $\mathcal{L}_\mathrm{TSA}$. As for semantics-level Shapley interaction modeling, the semantic alignment between the aggregated regions and textual phrases is learned, supervised by the fine-grained semantic alignment loss $\mathcal{L}_\mathrm{FSA}$. Combined with the two newly proposed losses, the full objective of fine-grained semantically aligned vision-language pre-training can be formulated as:
	
	\begin{equation}
		\mathcal{L} = \mathcal{L}_\mathrm{CMC} + \mathcal{L}_\mathrm{TSA} + \mathcal{L}_\mathrm{FSA}
	\end{equation}
	
	Such a new pre-training objective enforces the image encoder to capture semantic regions and establishes fine-grained semantic alignment between visual regions and textual phrases. During inference, it can be directly removed, rendering an efficient  and semantics-sensitive \emph{dual-encoder}.

	\subsection{Interpreting Fine-Grained Semantic Alignment as Game-Theoretic Interaction}
	\label{3.2}
	
	\subsubsection{Preliminaries}
	
	\textbf{Shapley Values.} The Shapley value~\cite{shapley1953value} is a classic game theory solution for the unbiased estimation of the importance or contribution of each player in a cooperative game. Considering a game with $\mathcal{N} = \{1, ..., n\}$ players, $\mathcal{S} \subseteq \mathcal{N}$ denotes a potential subset of players. A game $v(\cdot)$ is implemented as a function that maps each subset $\mathcal{S}$ of players to a score, modeling the outcome of a game when players in $\mathcal{S}$ participate in. Specifically, $v(\mathcal{N}) - v(\varnothing)$ denotes the contribution obtained by all players in the game. The Shapley value $\phi(i|\mathcal{N})$ for player $i$ is defined as the average marginal contribution of player $i$ to all possible coalitions $\mathcal{S}$ that are formed without $i$:

	\begin{equation}
		\phi(i|\mathcal{N}) = \sum_{\mathcal{S} \subseteq \mathcal{N}\setminus \{i\} } p(\mathcal{S}) [v(\mathcal{S}\cup\{i\}) - v(\mathcal{S})], \quad p(\mathcal{S}) = \frac{|\mathcal{S}|!(|\mathcal{N}| - |\mathcal{S}| - 1)!}{|\mathcal{N}|!} \label{e3}
	\end{equation}
	where $p(\mathcal{S})$ is the likelihood of $\mathcal{S}$ being sampled. The Shapley value has been proved to be the unique metric that satisfies the following axioms: \emph{Linearity}, \emph{Symmetry}, \emph{Dummy}, and \emph{Efficiency} \cite{weber1988probabilistic}. We summarize these axioms in Appendix B.
	
	\textbf{Shapley Interaction.} In the game theory, some players tend to form a coalition and always participate in the game together. The players in the coalition might interact or cooperate with each other, which brings additional contributions to the game. The Shapley interaction~\cite{grabisch1999axiomatic} measures this additional contributions brought by the coalition compared with the case when the players work individually. For a coalition $\mathcal{S}$, we consider $[\mathcal{S}]$ as a single hypothetical player, which is the union of the players in $\mathcal{S}$. Then, the reduced game is formed by removing the individual players in $\mathcal{S}$ from the game and adding $[\mathcal{S}]$ to the game. The Shapley value $\phi([\mathcal{S}]|\mathcal{N} \setminus \mathcal{S} \cup \{[\mathcal{S}]\})$ for player $[\mathcal{S}]$ can be computed using Equation~\ref{e3} over the reduced game. Similarly, we can obtain $\phi(i|\mathcal{N} \setminus \mathcal{S} \cup \{i\})$, where $i$ is the individual player in $S$. Finally, the Shapley interaction for coalition $\mathcal{S}$ is formulated as:
	
	\begin{equation} 
		\mathfrak{I}([\mathcal{S}]) = \phi([\mathcal{S}]|\mathcal{N} \setminus \mathcal{S} \cup \{[\mathcal{S}]\}) - \sum_{i \in \mathcal{S}} \phi(i|\mathcal{N} \setminus \mathcal{S} \cup \{i\}) \label{e4}
	\end{equation}
	
	In this way, $\mathfrak{I}([\mathcal{S}])$ reflects the interactions inside $\mathcal{S}$. The higher value of $\mathfrak{I}([\mathcal{S}])$ indicates that players in $\mathcal{S}$ cooperate closely with each other.

	\subsubsection{Token-Level Shapley Interaction Modeling}
	\vspace{-0.1cm}
	Due to inherent semantic unit mismatch between texts and images, it is ineffective to directly compute the alignment between words and pixels~(patches). A textual phrase usually refers to a specific image region, which is composed of multiple patches and represents a visual instance. Thus, we first introduce token-level Shapley interaction modeling to aggregate patches into semantic regions. 
	
	\textbf{Input Representations.} Given an image-text pair, the input image $I$ is sliced into patches and flattened. Followed by linear projection layer and position embeddings, we obtain patch token sequence $\mathcal{X}^I = \{\mathbf{x}^I_i\}_{i=1}^{L_1}$  with an additional \verb|[CLS_I]| token embedding. The input text $T$ is tokenized and embedded into word token sequence $\mathcal{X}^T = \{\mathbf{x}^T_i\}_{i=1}^{L_2}$, added with position embeddings. We also prepend a learnable special token \verb|[CLS_T]| to the word token sequence. Then, we adopt a dual-encoder structure to encode the patch token sequence and word token sequence, separately. On top of the image and text encoders, we obtain the representations of patch token sequence $\tilde{\mathcal{X}}^I = \{\tilde{\mathbf{x}}^I_i\}_{i=1}^{\tilde{L}_1}$ and word token sequence $\tilde{\mathcal{X}}^T = \{\tilde{\mathbf{x}}^T_i\}_{i=1}^{\tilde{L}_2}$. We take the learned representations of \verb|[CLS_I]| and \verb|[CLS_T]| tokens as the global representations for images and texts. And the global similarity of image-text pairs is measured by the cosine similarity between them.

	\textbf{Understanding Semantic Region via Shapley Interaction.} Supposing a set of patches represent a complete visual instance in an image, then they tend to have a strong Shapley interaction because they work jointly to form a visual instance, which contributes to the better similarity judgment with the text. From the game-theoretic view, we take patch tokens and word tokens as players $\mathcal{X} = \mathcal{X}^I \cup \mathcal{X}^T$, and the global similarity between images and texts as the game score $v_1(\cdot)$. To compute $v_1(\mathcal{S})$, we keep tokens in $\mathcal{S}$ and mask input tokens in $\mathcal{X} \setminus \mathcal{S}$ to zeros. Thus, the global similarity only considers the tokens in $\mathcal{S}$, which reflects the contribution of the tokens in $\mathcal{S}$ to the global similarity judgment.

	\textbf{Semantic Region Generation.} Inspired by \emph{YOLOv3}~\cite{2018YOLOv3}, we design a lightweight region generation module. It takes each patch token representation $\tilde{\mathbf{x}}^I_i$ as input and generates a bounding box prediction centered on $\tilde{\mathbf{x}}^I_i$, which corresponds to a visual region $\mathcal{R}_i = \{\mathbf{x}^I_{i, k} \}_{k=1}^{K_i}$ with ${K_i}$ patch tokens. The region generation module also predicts a confidence score $s(\mathcal{R}_i)$ for each region. We select the top-$M$ predictions as the semantic regions. Then, the Shapley interaction of $\mathcal{R}_i$ can be defined as:

	\begin{equation}
		\mathfrak{I}([\mathcal{R}_i]) = \phi([\mathcal{R}_i]|X \setminus \mathcal{R}_i \cup \{[\mathcal{R}_i]\}) - \sum_{\mathbf{x}_{i, k}^I \in \mathcal{R}_i} \phi(\mathbf{x}_{i, k}^I |\mathcal{X} \setminus \mathcal{R}_i \cup \{\mathbf{x}_{i,k}^I\}) 
	\end{equation}

	According to the Equation \ref{e3}, we can reformulate Shapley value into the form of expectation:

	\begin{equation}
		\phi([\mathcal{R}_i]|\mathcal{X} \setminus \mathcal{R}_i \cup \{[\mathcal{R}_i]\}) = \mathop{\mathbb{E}}\limits_{c}\{ \mathop{\mathbb{E}}\limits_{\mathcal{S} \subseteq \mathcal{X} \setminus \mathcal{R}_i  \atop |\mathcal{S}| = c} [v_1(\mathcal{S} \cup \mathcal{R}_i) - v_1(\mathcal{S})] \}
	\end{equation}
	where $c$ represents the coalition size. $\phi(\mathbf{x}_{i, k}^I |X \setminus \mathcal{R}_i \cup \{\mathbf{x}_{i, k}^I\})$ can be defined in a similar manner, and the Shapley interaction of $\mathcal{R}_i$ can be reformulated as (we provide the proof in Appendix C):

	\begin{equation}
		\mathfrak{I}([\mathcal{R}_i]) = \mathop{\mathbb{E}}\limits_{c}\{ \mathop{\mathbb{E}}\limits_{\mathcal{S} \subseteq \mathcal{X} \setminus \mathcal{R}_i  \atop |\mathcal{S}| = c} [v_1(\mathcal{S} \cup \mathcal{R}_i) - \sum_{\mathbf{x}_{i, k}^I \in \mathcal{R}_i} v_1(\mathcal{S} \cup \{\mathbf{x}_{i, k}^I\}) + (K - 1) v_1(\mathcal{S})]\}  \label{e7}
	\end{equation}

	Taking normalized $\mathfrak{I}'([R_i])$ as the soft supervision label, the token-based semantic aggregation loss is defined as cross-entropy loss:
	
	\begin{equation}
		\mathcal{L}_\mathrm{TSA} = -\frac{1}{M} \sum_{i=1}^{M} [ \mathfrak{I}'([\mathcal{R}_i]) \log(s(\mathcal{R}_i)) + (1 - \mathfrak{I}'([\mathcal{R}_i]) )\log(1 - s(\mathcal{R}_i)) ]
	\end{equation}
	which propagates gradients to the region generation module and image encoder to adjust bounding box predictions such that more accurate semantic regions can be captured.

	\subsubsection{Semantics-Level Shapley Interaction Modeling}\label{3.2.2}
	\vspace{-0.1cm}
	After obtaining the inferred semantic regions, we propose semantics-level Shapley interaction modeling to explicitly model the fine-grained semantic alignment between regions and phrases. We first define the fine-grained similarity score and then explain semantic alignment based on game theory. 
	
	We adopt \verb|Avg-Pooling| over learned patch representations in each $\mathcal{R}_i$ to obtain region representation $\mathbf{h}_i^I \in \mathbb{R}^d$. We employ an off-the-shelf constituency parser to extract phrases from text and obtain phrase representation $\mathbf{h}_i^T \in \mathbb{R}^d$ by \verb|Avg-Pooling|. Totally, we obtain $M$ regions $\mathcal{H}^I = \{\mathbf{h}^I_i\}_{i=1}^{M}$ and $N$ phrases $\mathcal{H}^T = \{\mathbf{h}^T_j\}_{j=1}^{N}$. And the alignment matrix can be defined as: $\mathbf{A} = [a_{ij}]^{M\times N}$, where $a_{ij}={\mathbf{h}^I_i}^\top \mathbf{h}^T_j$ represents the alignment score between $i$-th region and $j$-th phrase. Next, we apply softmax-normalization over each row of $\mathbf{A}$, obtaining $\tilde{\mathbf{A}}$. For the $i$-th region, we calculate its maximum alignment score as $\max_j \tilde{a}_{ij}$. Then, we use the average maximum alignment score over all regions as the fine-grained image-to-text similarity $p_1$. Similarly, we can obtain the fine-grained text-to-image similarity $p_2$, and the total fine-grained similarity score can be defined:  $p = (p_1 + p_2)/2$.
	
	\textbf{Understanding Semantic Alignment via Shapley Interaction.} If a region and a phrase have strong semantic correspondence, then they tend to cooperate with each other and contribute to the fine-grained similarity score. Thus, we can consider $\mathcal{H} = \mathcal{H}^I \cup \mathcal{H}^T$ as the players and the fine-grained similarity score $p$ as the game score $v_2(\cdot)$. The Shapley interaction of them can be formulated as:
	
	\begin{align}
		\mathfrak{I}([\mathcal{H}_{ij}]) &= \phi([\mathcal{H}_{ij}]|\mathcal{H} \setminus \mathcal{H}_{ij} \cup \{[\mathcal{H}_{ij}]\}) - \phi(\mathbf{h}^I_i|\mathcal{H} \setminus \mathcal{H}_{ij} \cup \{\mathbf{h}^I_i\}) - \phi(\mathbf{h}^T_j| \mathcal{H} \setminus \mathcal{H}_{ij} \cup \{\mathbf{h}^T_j\}) \\
		&
		= \mathop{\mathbb{E}}\limits_{c} \{\mathop{\mathbb{E}}\limits_{\mathcal{S} \subseteq \mathcal{H} \setminus \mathcal{H}_{ij}  \atop |\mathcal{S}| = c} [v_2(\mathcal{S} \cup \mathcal{H}_{ij}) - v_2(\mathcal{S} \cup \{\mathbf{h}^I_i\}) - v_2(\mathcal{S} \cup \{\mathbf{h}^T_j\}) +  v_2(\mathcal{S})] \} \label{e10}
	\end{align}
	where $[\mathcal{H}_{ij}]$ represents the single player formed by the coalition of $i$-th region and $j$-th phrase. Taking normalized $\mathfrak{I}'([\mathcal{H}_{ij}])$ as soft labels, the fine-grained semantic alignment loss can be defined as:
	
	\vspace{-0.15cm}
	\begin{equation}
		\mathcal{L}_\mathrm{FSA} =   -\frac{1}{MN} \sum_{i=1}^{M} \sum_{j=1}^{N} \mathfrak{I}'([\mathcal{H}_{ij}]) \log (\tilde{a}_{ij})
	\end{equation}
	
	\vspace{-0.15cm}
	\subsection{Uncertainty-Aware Neural Shapley Interaction Learning} 
	\label{3.3}
	
	According to Equation~\ref{e3} and Equation~\ref{e4}, computing exact the Shapley value is an NP-hard problem~\cite{matsui2001np}. Previous methods mainly apply sampling-based method~\cite{castro2009polynomial} to approximate it. While sampling-based approximation is unbiased, an accurate approximation requires thousands of model evaluations. To reduce the computational cost, we propose an uncertainty-aware neural Shapley interaction learning~(UNSIL) module to cooperate with the sampling-based method, rendering an efficient and effective hybrid strategy.
	
	Specifically, the sampling-based method~\cite{castro2009polynomial} estimates the expectation terms in Equation~\ref{e7} and Equation~\ref{e10} by sampling to compute the Shapley interaction. Inspired by noisy label learning~\cite{kendall2017uncertainties}, the UNSIL module learns to predict the Shapley interaction and the corresponding uncertainty $\sigma~\in~(0, 1)$. Intuitively, if the UNSIL module makes a prediction with low uncertainty, we can directly apply its prediction to $\mathcal{L}_\mathrm{TSA} $ and $\mathcal{L}_\mathrm{FSA}$, avoiding thousands of model evaluations. If the uncertainty is high, we then resort to the sampling-based method for a more accurate estimation.
	
	During training, the UNSIL module first predicts the target interaction with uncertainty $\sigma$. Then, we sample a value $\epsilon$ from a uniform distribution on $(0, 1)$. If $\epsilon \textgreater \sigma$, we directly use its prediction. If $\epsilon \leq \sigma$, we then use the sampling-based method to compute the Shapley Interaction and update the UNSIL module based on the sampling-based results. Note that, for the first few iterations, we employ the sampling-based method directly, and use its results to train the UNSIL module. 

	Let ${\mathfrak{I}}^*$ and $\hat{\mathfrak{I}}$ denote the results from the sampling-based method and UNSIL module, respectively. Taking ${\mathfrak{I}}^*$ as the ground-truth, the UNSIL module  is trained by:
	
	\vspace{-0.1cm}
	\begin{equation}
		\mathcal{L}_\mathrm{UNSIL} = \frac{1}{\beta_1\sigma} \mathcal{L}_\mathrm{MSE}( \hat{\mathfrak{I}},  {\mathfrak{I}}^*) + \beta_2 \sigma
	\end{equation}

	where the first term is mean squared error $\mathcal{L}_\mathrm{MSE}$ weighted by the uncertainty, the second term serves as a regularization term for the prediction uncertainty, and $\beta$ is the weight hyper-parameter. The UNSIL module implicitly learns the uncertainty from the regression loss function. We discuss the implementation details of the UNSIL module in Section~\ref{4.4} and Appendix D.

	\section{Experiments}
	
	\vspace{-0.2cm}
	\subsection{Pre-training Details}\label{4.1}
	\vspace{-0.2cm}
	As sufficient data is a prerequisite for vision-language pre-training, we construct a dataset with 240M image-text pairs from the Internet. We implement the image encoder by Swin-L~\cite{liu2021Swin} and the text encoder by BERT-Small~\cite{devlin2018bert}. The input images are resized to $224 \times 224$ and the input texts are tokenized by WordPiece with a maximum length of 60. We pre-train the model for 20 epochs using a batch size of 512 on 128 NVIDIA V100 GPUs. We utilize AdamW~\cite{loshchilov2017decoupled} optimizer with a learning rate of $2\times10^{-4}$ and a weight decay of 0.01. More pre-training and evaluation details are provided in Appendix D, E. We also analyze the image encoder and training efficiency in Appendix G, J.
	
	\begin{table}[!t]
		\caption{Results (\%) of zero-shot image-text retrieval on Flickr30K and MSCOCO datasets.}
		\label{t_retrieval}
		\resizebox{\textwidth}{!}{
			\centering
			\begin{tabular}{ ll| cccccc |cccccc}
				\hline
				\multicolumn{2}{l|}{\multirow{2}*{}} 
				&\multicolumn{6}{c|}{\textsl{Flickr30K}} 
				&\multicolumn{6}{c}{\textsl{MSCOCO}}
				\\

				\multicolumn{2}{l|}{} &\multicolumn{3}{c}{\textsl{image-to-text}} &\multicolumn{3}{c|}{\textsl{text-to-image}} &\multicolumn{3}{c}{\textsl{image-to-text}} &\multicolumn{3}{c}{\textsl{text-to-image}} \\
				\hline
				\multicolumn{2}{l|}{} &R@1 &R@5 &R@10 &R@1 &R@5 &R@10 &R@1 &R@5 &R@10 &R@1 &R@5 &R@10 \\
				
				\multicolumn{2}{l|}{ImageBERT} &70.7 &90.2 &94.0 &54.3 &79.6 &87.5 &44.0 &71.2 &80.4 &32.3 &59.0 &70.2 \\
				\multicolumn{2}{l|}{UNITER}  &83.6 &95.7 &97.7  &68.7 &89.2 &93.9  &- &-  &-  &- &-  &-  \\
				\multicolumn{2}{l|}{CLIP}  &88.0 &98.7  &99.4  &68.7  &90.6 &95.2  &58.4 &81.5 &88.1  &37.8 &62.4 &72.2 \\
				\multicolumn{2}{l|}{ALIGN}  &88.6  &98.7  &99.7  &75.7 &93.8 &\textbf{96.8}  &58.6 &83.0 &89.7  &45.6 &69.8 &78.6  \\
				\multicolumn{2}{l|}{FILIP}  &89.8 &99.2 &99.8  &75.0 &93.4  &96.3  &61.3 &84.3 &90.4  &45.9 &70.6 &79.3  \\
				
				\multicolumn{2}{l|}{\textbf{LOUPE}}  &\textbf{90.5}  &\textbf{99.5}  &\textbf{99.8}  &\textbf{76.3}  &\textbf{93.9}  &96.7  &\textbf{62.3}  &\textbf{85.1} &\textbf{91.2} &\textbf{50.1}  &\textbf{75.4} &\textbf{83.3}  \\
				
				\hline    
				
			\end{tabular}
			
		}
		\vspace{-0.3cm}
	\end{table}

	\begin{table}[!t]
	\caption{Top-1 accuracy (\%) of zero-shot image classification over 11 datasets.}
	\label{zero_classification}
		\centering
		\begin{tabular}{ l| ccccccccccc|c}
			\hline
			&\rotatebox{90}{CIFAR10}  &\rotatebox{90}{Food101} &\rotatebox{90}{StanfordCars} &\rotatebox{90}{SUN397} &\rotatebox{90}{Flowers102} &\rotatebox{90}{Country211} &\rotatebox{90}{FER2013} &\rotatebox{90}{Aircrafts} &\rotatebox{90}{OxfordPets} &\rotatebox{90}{Caltech101} &\rotatebox{90}{ImageNet} &\rotatebox{90}{Average}
			\\

			\hline
			
			CLIP     &\textbf{96.2}   &92.9   &77.3    &67.7   &78.7   &34.9 &\textbf{57.7}   &36.1   &93.5 &92.6   &75.3  &73.0   \\

			\textbf{LOUPE}     &95.9   &\textbf{94.3}   &\textbf{79.9}    &\textbf{69.8}   &\textbf{87.4}   &\textbf{37.8}  &53.3   &\textbf{54.9}   &\textbf{94.1}  &\textbf{93.9}   &\textbf{76.1}  &\textbf{76.1}  \\
			
			\hline    
			
		\end{tabular}
		
	\vspace{-0.3cm}
\end{table}
	
	\subsection{Zero-Shot Image-Text Retrieval}
	We compare LOUPE on the widely used MSCOCO~\cite{lin2014microsoft} and Flickr30K~\cite{plummer2015flickr30k} datasets. First, the results in Table \ref{t_retrieval} show that LOUPE achieves new state-of-the-art zero-shot performance on most metrics of the two datasets, demonstrating the stronger generalizability of our pre-training framework. Second, while previous works mainly pre-train on larger datasets~(CLIP 400M, ALIGN 1800M, FILIP 340M), LOUPE still achieves superior performance using less training data~(240M). Third, compared with FILIP which directly computes token-wise similarity, our model captures semantic alignment between visual regions and textual phrases, which is more semantically meaningful. For text-to-image retrieval on MSCOCO, LOUPE significantly outperforms FILIP by 4.2\% on recall@1.

	\vspace{-0.1cm}
	\subsection{Zero-Shot Image Classification}
	\vspace{-0.15cm}
	In this section, we evaluate LOUPE on the zero-shot image classification task. We compare LOUPE with CLIP on 11 downstream classification datasets, following the same evaluation setting as CLIP~\cite{radford2021learning}. Table~\ref{zero_classification} summarizes the results. As shown in Table~\ref{zero_classification}, our LOUPE outperforms CLIP with average improvement of 3.1\%. Notably, on ImageNet, the largest dataset among 11 datasets, our LOUPE surpasses CLIP by 0.8\%. Also, we observe that LOUPE achieves substantial performance gains on several fine-grained image classification datasets (i.e., Flowers102 and Aircrafts). It demonstrates the superiority of our LOUPE on fine-grained semantics understanding.
	
	We also evaluate the linear probing performance of our LOUPE on image classification. The detailed results can be found in Appendix I.

	\begin{table}[!t]
		\caption{Without fine-tuning, zero-shot transfer performance on object detection and visual grounding.}
		\label{t_detection_grounding}
		\resizebox{\textwidth}{!}{
			\centering
			\begin{tabular}{ ll| cc|cc| ccc| ccc}
				\hline
				\multicolumn{2}{l|}{\multirow{2}*{}} 
				&\multicolumn{2}{c|}{\textsl{COCO}} 
				&\multicolumn{2}{c|}{\textsl{PASCAL VOC}} 
				&\multicolumn{3}{c|}{\textsl{RefCOCO}} 
				&\multicolumn{3}{c}{\textsl{RefCOCO+}}
				\\

				\multicolumn{2}{l|}{}   &mAP@0.3 &mAP@0.5 &mAP@0.3  &mAP@0.5 &val  &testA &testB  &val  &testA &testB   \\
				\hline
				\multicolumn{2}{l|}{CLIP + Pixel-Wise}      &8.5 &4.5  &18.2   &7.3  &6.7   &6.2 &5.8  &6.1   &7.0 &5.7  \\
				\multicolumn{2}{l|}{CLIP + K-Means}          &6.4 &1.9  &11.7    &4.8     &2.1    &2.3 &1.7   &1.7 &2.0 &2.8  \\
				\multicolumn{2}{l|}{CLIP + Grad-CAM}      &7.1  &3.2  &19.1  &8.2  &5.5 &5.2 &4.8 &4.4   &5.6  &4.9  \\
				\multicolumn{2}{l|}{AdaptCLIP}                    &14.9  &6.6   &28.7  &12.9   &16.7   &18.4   &18.0   &17.5   &18.9  &19.6  \\
				\multicolumn{2}{l|}{\textbf{LOUPE}}           &\textbf{25.3}  &\textbf{12.1}   &\textbf{30.3} &\textbf{19.5} &\textbf{25.2} &\textbf{26.8} &\textbf{24.5}  &\textbf{22.9}   &\textbf{23.3} &\textbf{23.6}  \\
				\hline    
				
			\end{tabular}
			
		}
		\vspace{-0.2cm}
	\end{table}

	\begin{figure}[!t]
		\centering
		\includegraphics[width=\textwidth]{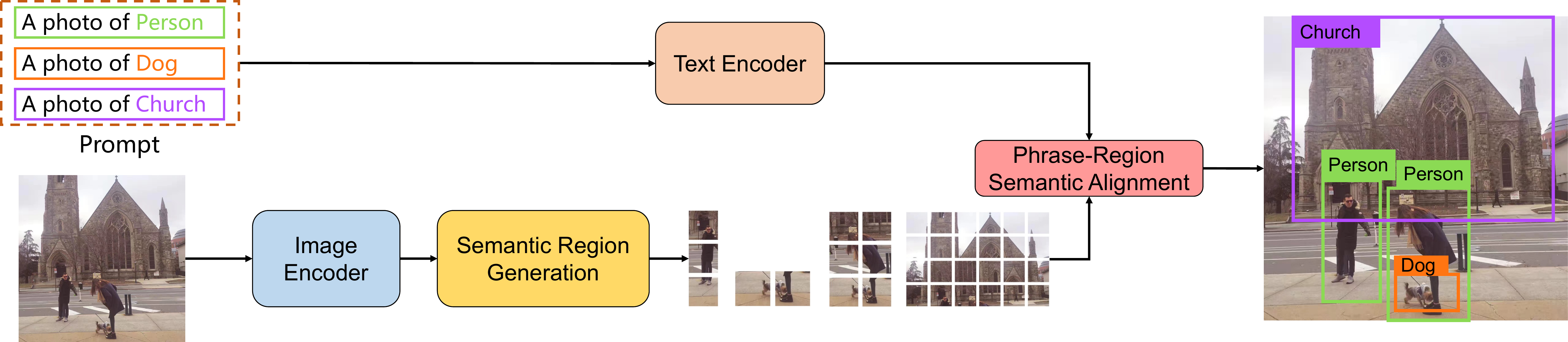}
		\vspace{-0.4cm} 
		\caption{An example of  LOUPE zero-shot transferring to object detection using prompt templates.}
		\label{detection}
		\vspace{-0.4cm} 
	\end{figure}

	\vspace{-0.2cm}
	\subsection{Zero-Shot Transfer  to Object Detection and Visual Grounding}\label{4.3}
	\vspace{-0.15cm}

	To answer whether our model has learned fine-grained semantics, we further evaluate LOUPE on object detection~\cite{ren2015faster} and visual grounding~\cite{yu2016modeling}, which require more fine-grained semantic understanding ability to identify specific visual regions in images according to the object labels or referring expressions.
	Visual grounding can be seen as generalized object detection, where the pre-defined class labels are replaced by language referring expression sentences. As LOUPE can generate a set of semantic regions that are aligned with textual phrases, it can 
	be easily applied to object detection and visual grounding without structure modification. For visual grounding, we take referring expressions as input text. For object detection, as illustrated in Figure~\ref{detection}, we use prompt to expand detection labels to input text. Then, we encode input text by the learned text encoder, and these tasks can be completed by measuring the similarity between candidate region representations and text representations.
	
	For comparison, we zero-shot transfer CLIP (ViT-L/14) to object detection and visual grounding by applying several non-parametric approaches on the spatial feature maps of CLIP. We also compare with AdaptCLIP~\cite{li2022adapting}, which is a concurrently unpublished method that leverages  classic super-pixel~(SLIC~\cite{achanta2012slic}) and bounding box proposal (selective search~\cite{uijlings2013selective}) methods to zero-shot transfer CLIP to phrase localization. We use its public official implementations to get the experiment results. For object detection, we evaluate their mean Average Precision (mAP) at IoU thresholds of $\{0.3, 0.5\}$ on COCO~\cite{lin2014microsoft} (65 classes) and PASCAL VOC~\cite{everingham2008pascal} (20 classes). For visual grounding, we evaluate their top-1 accuracy at IoU thresholds of 0.5 on RefCOCO~\cite{yu2016modeling} and RefCOCO+~\cite{yu2016modeling}. The experiment details of CLIP variants  and LOUPE  are provided in Appendix E.

	Table \ref{t_detection_grounding} summarizes the results. \textbf{1)} Overall, LOUPE outperforms all CLIP variants by a large margin. The significantly higher performance illustrates the stronger zero-shot transfer ability of our fine-grained semantically aligned pre-training paradigm. \textbf{2)} Second, all CLIP variants rely on pre-processing steps on CLIP's feature map (\textsl{e.g.,} AdaptCLIP first uses SLIC to group pixels and then uses selective search to generate a large number of proposals), which is time-consuming. In contrast, our method directly predicts the semantic regions based on the  patch token representations. \textbf{3)} Third, the consistently competitive performance across four benchmarks validates that LOUPE can learn fine-grained semantics from raw text supervision. LOUPE demonstrates a promising alternative, that is, learning fine-grained semantics from large-scale raw image-text pairs, which are easily available and contain a broader set of visual concepts.
	
	As time-consuming human annotations are unscalable for massive object classes in the real world, some recent works~\cite{bansal2018zero, rahman2020improved} target at training object detectors with annotations on base object classes to generalize to  the remaining object classes of the same dataset. The latest works~\cite{gu2021open, zareian2021open} leverage the generalizability of vision-language pre-training models to further improve the zero-shot performance on novel classes. However, these zero-shot approaches still require bounding box annotations on base classes for task-specific supervised learning. 
	In contrast, our LOUPE is trained on large-scale raw image-text pairs, which are already accessible on the Internet and contain more diverse semantics.

	\begin{table}[!t]
		\caption{Ablation study of each component across three tasks.}
		\label{t_ablation}
		\resizebox{\textwidth}{!}{
			\centering
			\begin{tabular}{ lll| cc| cc |ccc|c}
				\hline
				&\multicolumn{2}{l|}{\multirow{2}*{}} 
				&\multicolumn{2}{c|}{\textsl{MSCOCO}} 
				&\multicolumn{2}{c|}{\textsl{COCO}}
				&\multicolumn{3}{c|}{\textsl{RefCOCO}}
				&Training Time
				\\

				&\multicolumn{2}{l|}{}  &I2T  &T2I &mAP@0.3 &mAP@0.5  &val  &testA &testB &(sec/iter) \\
				\hline
				1&\multicolumn{2}{l|}{Backbone}  &31.0  &24.8   &3.8   &1.0   &1.3  &0.9  &0.8  &1.17  \\
				2&\multicolumn{2}{l|}{Backbone + $\mathcal{L}_\mathrm{TSA}$}  &32.4  &26.2   &7.6   &3.3  &1.8 &2.0 &2.6 &8.38  \\
				3&\multicolumn{2}{l|}{Backbone + $\mathcal{L}_\mathrm{TSA}$ + $\mathcal{L}_\mathrm{FSA}$}  &33.5   &28.3  &9.4  &5.9 &4.1  &4.6 &4.3  &9.90  \\
				4&\multicolumn{2}{l|}{Backbone + $\mathcal{L}_\mathrm{TSA}$ + $\mathcal{L}_\mathrm{FSA}$ + UNSIL}  &33.3  &28.1   &9.0  &5.6  &4.5 &4.9 &4.4 &1.93  \\
				\hline    
				
			\end{tabular}
		}
		\vspace{-0.2cm}
	\end{table}
	
	\begin{figure}[!t]
		\centering
		\includegraphics[width=\textwidth]{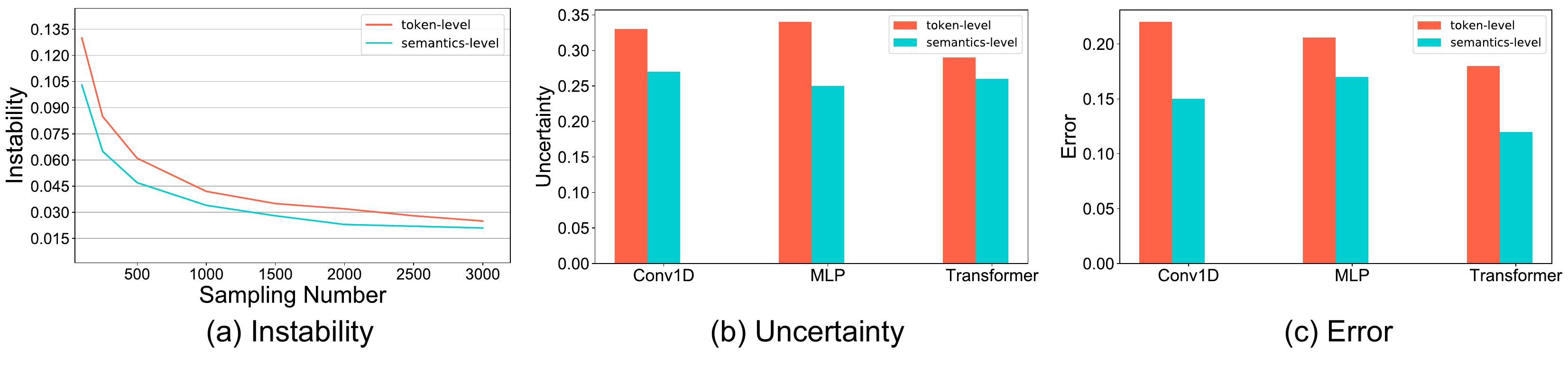}
		\vspace{-0.7cm}
		\caption{(a) Instability of the Shapley interaction approximation with respect to different sampling numbers. (b, c) Uncertainty and error of the UNSIL module with different structures.}
		\label{ablation_shapley}
		\vspace{-0.5cm}
	\end{figure}

	\vspace{-0.2cm}
	\subsection{Ablation Study} \label{4.4}
	\vspace{-0.2cm}
	\textbf{Effectiveness of Individual Components.} In this section, we investigate the effectiveness of each component in Table~\ref{t_ablation}. Given the costly training time, all ablation studies are based on a relatively small dataset (Conceptual Captions 3M~\cite{2018Conceptual}). We start with the backbone model that consists of a dual-encoder trained by cross-modal contrastive loss. We then gradually add token-level Shapley interaction modeling supervision $\mathcal{L}_\mathrm{TSA}$~(Row 2), semantics-level Shapley interaction modeling supervision $\mathcal{L}_\mathrm{FSA}$~(Row 3), and UNSIL module~(Row 4). For Row 2 and Row 3, the Shapley interaction is only computed by the sampling-based method. The results in Table~\ref{t_ablation} show that both $\mathcal{L}_\mathrm{TSA}$ and $\mathcal{L}_\mathrm{FSA}$ bring significant improvement for all tasks. We observe that $\mathcal{L}_\mathrm{TSA}$ boosts a 3.8\% improvement on object detection. And the improved fine-grained visual semantic understanding further facilitates the cross-modal retrieval performance~(+1.4\%). The  semantics-level Shapley interaction modeling further improves the performance on all tasks by modeling the semantic alignment between visual regions and textual phrases. Comparing Row 3 and Row 4, we observe that the UNSIL module  maintains the estimation accuracy while avoiding intensive computations. The averaged training time is reduced from 9.90 seconds per iteration to 1.93 seconds per iteration. 
	
	
	\textbf{Accuracy of the Shapley Interaction Learning.} Since we use the sampling-based method~\cite{castro2009polynomial} to compute the Shapley Interaction and train the UNSIL module, we conduct a study to evaluate the accuracy of the sampling-based method and the error of the UNSIL module. As~\cite{zhang2020interpreting}, we compute the interaction multiple times and measure the instability of them. A lower instability means that we obtain similar interactions from different sampling processes. It indicates a high accuracy. Specifically, the instability is defined as $\frac{\mathbb{E}_{u, v: u \neq v}|\mathfrak{I}_u - \mathfrak{I}_v|}{\mathbb{E}_w |\mathfrak{I}_w|}$, where $\mathfrak{I}_w$ denotes the interaction computed in the $w$-th time. We average the instability values over Shapley interaction of 100 image-text pairs. We report the average instability values with respect to different sampling numbers. As shown in Figure \ref{ablation_shapley} (a), the instability decreases along with the increase of the sampling number. When the sampling number is larger than 500, the approximated Shapley interaction is stable enough with instability less than 0.06. 
	Further, we attempt different models (\textsl{i.e., Conv1D, 3-Layer MLP + Attention, 3-Layer Transformer}) to implement the UNSIL module (please see Appendix D for more details). We test them on 1000 samples and report their mean uncertainty and relative error in Figure \ref{ablation_shapley} (b) and (c). 
	We observe that \textsl{MLP + Attention} is good enough to predict the interaction with lower complexity. Thus, we implement the UNSIL module by \textsl{MLP + Attention}. 
	
	\begin{figure}[!t]
		\centering
		\includegraphics[width=\textwidth]{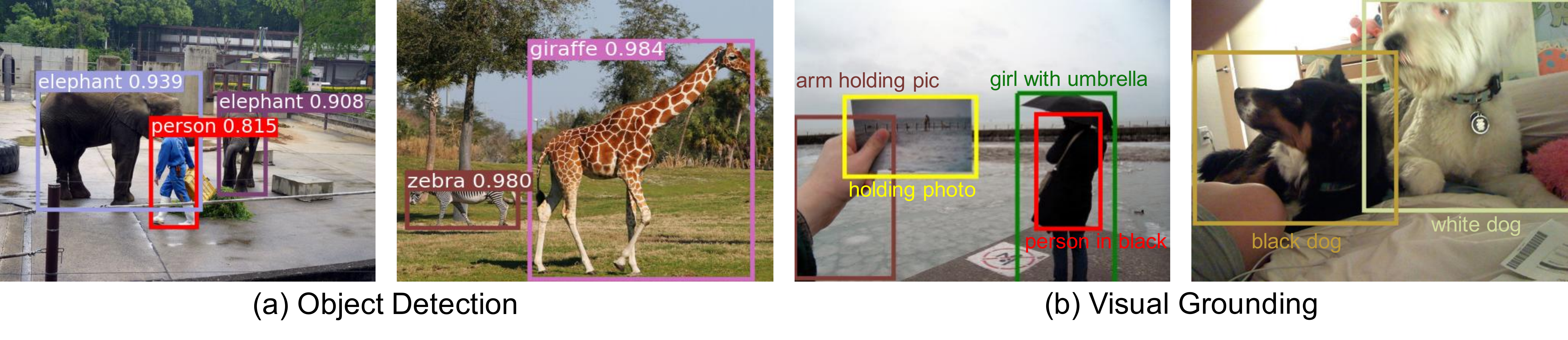}
		\vspace{-0.65cm} 
		\caption{Qualitative examples of object detection on COCO and  visual grounding on RefCOCO+.}
		\label{qualitative}
	\end{figure}
	
	\begin{figure}[!t]
		\centering
		\includegraphics[width=\textwidth]{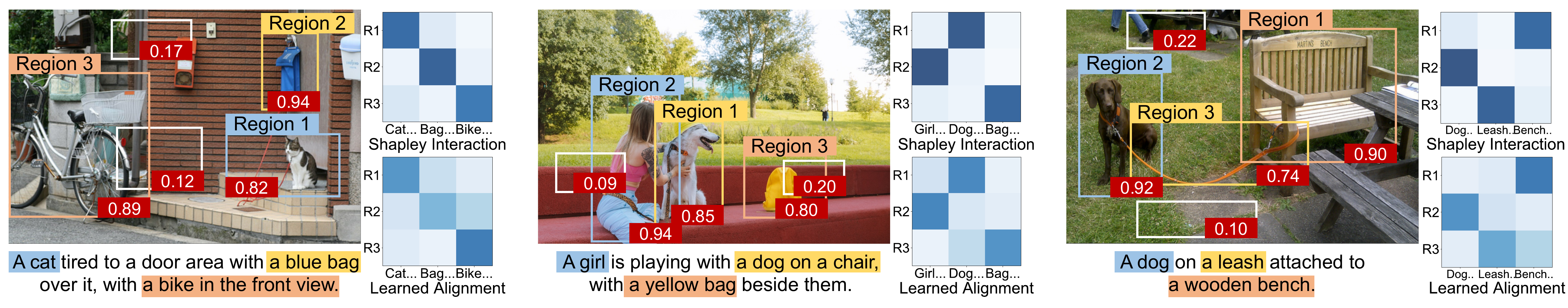}
		\vspace{-0.55cm}
		\caption{Visualization of learned fine-grained semantic alignment and corresponding Shapley interaction values. The values in the red boxes represent the Shapley interaction of regions.}
		\label{vis_shapley}
		\vspace{-0.2cm} 
	\end{figure}
	
	\vspace{-0.2cm}
	\subsection{Qualitative Analysis}
	\vspace{-0.1cm}
	\textbf{Qualitative Examples.} As shown in Figure~\ref{qualitative}, LOUPE successfully captures the regions that correspond to the detected objects, and grounds the referring expressions onto the referred regions.
	
	\textbf{Visualization of Learned Fine-Grained Semantic Alignment.} In Figure~\ref{vis_shapley}, we visualize some key semantic regions and corresponding alignment matrices inferred by LOUPE. We present the regions with top-3 confidence~(Region 1 -- 3) and two randomly sampled regions~(white boxes). The red boxes at the bottom of bounding boxes indicate their normalized token-level Shapley interaction values. Comparing their Shapley interaction values, we observe that the token-level Shapley interaction successfully distinguishes semantic regions from randomly sampled regions. The semantically meaningful regions tend to have stronger interaction. It indicates that token-level Shapley interaction can provide correct supervision for semantic region generation. Further, we show the alignment matrices inferred by semantics-level Shapley interaction and LOUPE, respectively. As shown in the right case of Figure~\ref{vis_shapley}, LOUPE successfully recognizes the leash region \includegraphics[height=0.1in, width=0.3in]{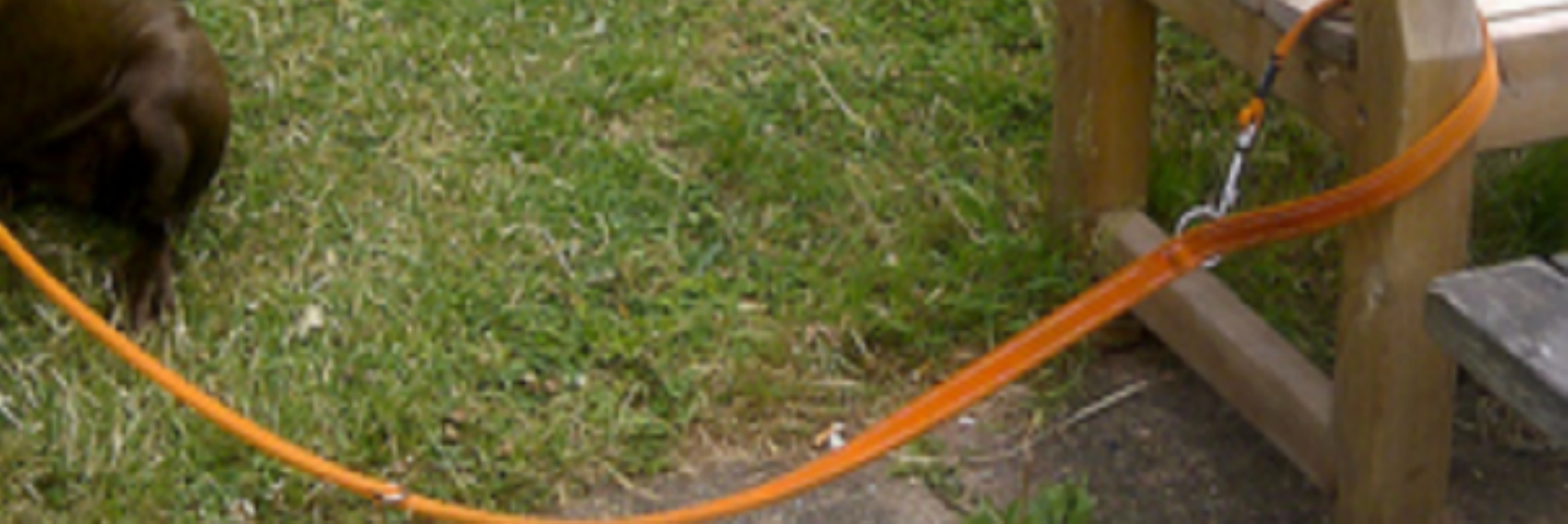} and aligns it with the ``a leash'' phrase. Note that existing object detection datasets do not contain the ``leash'' category. 
	
	\section{Conclusion}\label{s5}
	\vspace{-0.1cm}
	This paper introduces a novel vision-language pre-training framework, \textbf{\textsl{LOUPE}}\includegraphics[height=0.106in, width=0.11in]{loupe.pdf}, which models the fine-grained semantic alignment between visual regions and textual phrases by game-theoretic interactions. To efficiently compute the interactions, we further propose an uncertainty-aware neural Shapley interaction learning module. Comprehensive experiments show that LOUPE achieves new state-of-the-art on image-text retrieval datasets and can transfer to object detection and visual grounding in a zero-shot manner. This work demonstrates a new promising direction of learning fine-grained semantics from large-scale raw image-text data.
	
	\textbf{Limitations.} 1) The phrases are extracted by off-the-shelf constituency parsers, whose predictions might not be completely accurate. 2) The web data might inevitably contain mismatched image-text pairs, leading to noisy supervision.
	
	\textbf{Social Impacts.} Our model is trained on noisy data from the Internet, which may contain unsuitable images, violent text, or private information. Thus, additional analysis of the data is necessary. Further, the use of our model for privacy surveillance or other nefarious purposes should be prohibited.
	
	\textbf{Acknowledgment.} This work has been supported in part by the National Key Research and Development Program of China (2018AAA0101900), Zhejiang NSF (LR21F020004), Key Research and Development Program of Zhejiang Province, China (No. 2021C01013), Chinese Knowledge Center of Engineering Science and Technology (CKCEST). We thank all the reviewers for their valuable comments.

		\appendix

	\section{Appendix Overview}
	In this appendix, we present:
	
	\begin{itemize}[leftmargin=0.8cm]
		
		\item Axiomatic Properties of Shapley Value (Section~\ref{a1}).
		
		\item Proofs of Equation 7 and Equation 10 (Section~\ref{a2}).
		
		\item Hyperparameters and Implementation Details (Section~\ref{a3}).
		
		\item Pre-Training and Evaluation Details (Section~\ref{a4}).
		
		\item More Experiment Results on Downstream Vision-Language Generation Task (Section~\ref{a5}).
		
		\item Further Analysis on the Image Encoder (Section~\ref{a6}).
		
		\item More Qualitative Examples on Object Detection and Visual Grounding (Section~\ref{a7}).
		
		\item Linear Probing Evaluation (Section~\ref{a9}).
		
		\item Training Efficiency Discussion (Section~\ref{a10}).
		
		\item Detailed Discussion with Some Related Works (Section~\ref{a11}).
		
	\end{itemize}

	\section{Axiomatic Properties of Shapley Value}\label{a1}
	In this section, we mainly introduce the axiomatic properties of Shapley value. \textsl{Weber et al.}~\cite{weber1988probabilistic} have proved that Shapley value is the unique metric that satisfies the following axioms: \emph{Linearity}, \emph{Symmetry}, \emph{Dummy}, and \emph{Efficiency}.
	
	\textbf{Linearity Axiom.} If two independent games $u$ and $v$ can be linearly merged into one game $w(\mathcal{S}) = u(\mathcal{S}) + v(\mathcal{S})$, then the Shapley value of each player $i \in \mathcal{N}$ in the new game $w$ is the sum of Shapley values of the player $i$ in the game $u$ and $v$, which can be formulated as:
	
	\begin{equation}
		\phi_w(i|\mathcal{N}) = \phi_u(i|\mathcal{N}) + \phi_v(i|\mathcal{N})
	\end{equation}

	\textbf{Symmetry Axiom.} Considering two players $i$ and $j$ in a game $v$, if they satisfy:
	\begin{equation}
		\forall \mathcal{S} \in \mathcal{N} \setminus \{i, j\}, v(\mathcal{S} \cup \{i\}) = v(\mathcal{S} \cup \{j\})
	\end{equation}
	then $\phi_v(i|\mathcal{N}) = \phi_v(j|\mathcal{N})$.
	
	\textbf{Dummy Axiom.} The dummy player is defined as the player that has no interaction with other players. Formally, if a player $i$ in a game $v$ satisfies:
	\begin{equation}
		\forall \mathcal{S} \in \mathcal{N} \setminus \{i\}, v(\mathcal{S} \cup \{i\}) = v(\mathcal{S}) + v(\{i\})
	\end{equation}
	then this player is defined as the dummy player. In this way, the dummy player $i$ has no interaction with other players, \textsl{i.e.,} $v(\{i\}) = \phi_v(i|\mathcal{N})$.

	\textbf{Efficiency Axiom.} The efficiency axiom ensures that the overall reward can be assigned to all players, which can be formulated as:
	\begin{equation}
		\sum_{i \in \mathcal{N}} \phi_v(i) = v(\mathcal{N}) - v(\varnothing)
	\end{equation}

	\section{Proofs of Equation 7 and Equation 10}\label{a2}
	In this section, we provide detailed proofs for Equation~7 in Section~3.2.2 and Equation~10 in Section~3.2.3. 
	
	We first provide proof for Equation~7. The token-level Shapley interaction for $\mathcal{R}_i$ can be decomposed as follows:
	\begin{align}
		\mathfrak{I}([\mathcal{R}_i]) &= \phi([\mathcal{R}_i]|X \setminus \mathcal{R}_i \cup \{[\mathcal{R}_i]\}) - \sum_{\mathbf{x}_{i, k}^I \in \mathcal{R}_i} \phi(\mathbf{x}_{i, k}^I |\mathcal{X} \setminus \mathcal{R}_i \cup \{\mathbf{x}_{i,k}^I\}) \\
		&= \mathop{\mathbb{E}}\limits_{c}\{ \mathop{\mathbb{E}}\limits_{\mathcal{S} \subseteq \mathcal{X} \setminus \mathcal{R}_i  \atop |\mathcal{S}| = c} [v_1(\mathcal{S} \cup \mathcal{R}_i) - v_1(\mathcal{S})]\}  -  \sum_{\mathbf{x}_{i, k}^I \in \mathcal{R}_i} \mathop{\mathbb{E}}\limits_{c}\{ \mathop{\mathbb{E}}\limits_{\mathcal{S} \subseteq \mathcal{X} \setminus \mathcal{R}_i  \atop |\mathcal{S}| = c}[ v_1(\mathcal{S} \cup \{\mathbf{x}_{i, k}^I\}) - v_1(\mathcal{S})]\}\\
		&= \mathop{\mathbb{E}}\limits_{c}\{ \mathop{\mathbb{E}}\limits_{\mathcal{S} \subseteq \mathcal{X} \setminus \mathcal{R}_i  \atop |\mathcal{S}| = c} [v_1(\mathcal{S} \cup \mathcal{R}_i) - v_1(\mathcal{S})]\}  -   \mathop{\mathbb{E}}\limits_{c}\{ \mathop{\mathbb{E}}\limits_{\mathcal{S} \subseteq \mathcal{X} \setminus \mathcal{R}_i  \atop |\mathcal{S}| = c}[ \sum_{\mathbf{x}_{i, k}^I \in \mathcal{R}_i} (v_1(\mathcal{S} \cup \{\mathbf{x}_{i, k}^I\}) - v_1(\mathcal{S})\ )]\}\\
		&= \mathop{\mathbb{E}}\limits_{c}\{ \mathop{\mathbb{E}}\limits_{\mathcal{S} \subseteq \mathcal{X} \setminus \mathcal{R}_i  \atop |\mathcal{S}| = c} [v_1(\mathcal{S} \cup \mathcal{R}_i) - v_1(\mathcal{S}) - \sum_{\mathbf{x}_{i, k}^I \in \mathcal{R}_i}(v_1(\mathcal{S} \cup \{\mathbf{x}_{i, k}^I\}) - v_1(\mathcal{S})\ )\ ]\ \}\\
		&= \mathop{\mathbb{E}}\limits_{c}\{ \mathop{\mathbb{E}}\limits_{\mathcal{S} \subseteq \mathcal{X} \setminus \mathcal{R}_i  \atop |\mathcal{S}| = c} [v_1(\mathcal{S} \cup \mathcal{R}_i) - v_1(\mathcal{S}) - \sum_{\mathbf{x}_{i, k}^I \in \mathcal{R}_i} v_1(\mathcal{S} \cup \{\mathbf{x}_{i, k}^I\}) + \sum_{\mathbf{x}_{i, k}^I \in \mathcal{R}_i} v_1(\mathcal{S})\ ]\ \}\\
		&=  \mathop{\mathbb{E}}\limits_{c}\{ \mathop{\mathbb{E}}\limits_{\mathcal{S} \subseteq \mathcal{X} \setminus \mathcal{R}_i  \atop |\mathcal{S}| = c} [v_1(\mathcal{S} \cup \mathcal{R}_i) - \sum_{\mathbf{x}_{i, k}^I \in \mathcal{R}_i} v_1(\mathcal{S} \cup \{\mathbf{x}_{i, k}^I\}) + (K - 1) v_1(\mathcal{S}\ )\ ]\ \} 
	\end{align}
	
	We then provide proof for Equation~10. The semantics-level Shapley interaction between region $i$ and phrase $j$ can be decomposed as follows:
	
	\begin{align}
		\mathfrak{I}([\mathcal{H}_{ij}]) &= \phi([\mathcal{H}_{ij}]|\mathcal{H} \setminus \mathcal{H}_{ij} \cup \{[\mathcal{H}_{ij}]\}) - \phi(\mathbf{h}^I_i|\mathcal{H} \setminus \mathcal{H}_{ij} \cup \{\mathbf{h}^I_i\}) - \phi(\mathbf{h}^T_j| \mathcal{H} \setminus \mathcal{H}_{ij} \cup \{\mathbf{h}^T_j\}) \\
		&= \mathop{\mathbb{E}}\limits_{c} \{\mathop{\mathbb{E}}\limits_{\mathcal{S} \subseteq \mathcal{H} \setminus \mathcal{H}_{ij}  \atop |\mathcal{S}| = c} [v_2(\mathcal{S} \cup \mathcal{H}_{ij}) - v_2(\mathcal{S})] \} - \mathop{\mathbb{E}}\limits_{c} \{\mathop{\mathbb{E}}\limits_{\mathcal{S} \subseteq \mathcal{H} \setminus \mathcal{H}_{ij}  \atop |\mathcal{S}| = c} [v_2(\mathcal{S} \cup \{\mathbf{h}_i^I\}) - v_2(\mathcal{S})] \}\\
		& - \mathop{\mathbb{E}}\limits_{c} \{\mathop{\mathbb{E}}\limits_{\mathcal{S} \subseteq \mathcal{H} \setminus \mathcal{H}_{ij}  \atop |\mathcal{S}| = c} [v_2(\mathcal{S} \cup \{\mathbf{h}_j^T\}) - v_2(\mathcal{S})] \}\\
		&= \mathop{\mathbb{E}}\limits_{c} \{\mathop{\mathbb{E}}\limits_{\mathcal{S} \subseteq \mathcal{H} \setminus \mathcal{H}_{ij}  \atop |\mathcal{S}| = c} [v_2(\mathcal{S} \cup \mathcal{H}_{ij}) - v_2(\mathcal{S} \cup \{\mathbf{h}^I_i\}) - v_2(\mathcal{S} \cup \{\mathbf{h}^T_j\}) +  v_2(\mathcal{S})\ ]\  \}
	\end{align}

	\begin{table}[!t]
		\caption{A summary of various hyperparameters in LOUPE.}
		\label{hyper}
			\centering
			\begin{tabular}{ l r}
				\hline
				\textbf{Hyperparameter} &\textbf{Value} \\
				\hline
				\multicolumn{2}{c}{\textsl{Image Encoder - Swin-L}} \\
				\hline
				input image size                        &$224\times224$\\
				\hdashline
				stage 1 -  patch size                 &$4\times4$\\
				stage 1 -  hidden size               &192 \\
				stage 1 -  window size              &$7\times7$\\
				stage 1 -  number of heads     &6 \\
				\hdashline
				stage 2 -  patch size                 &$8\times8$\\
				stage 2 -  hidden size               &384 \\
				stage 2 -  window size              &$7\times7$\\
				stage 2 -  number of heads     &12 \\
				\hdashline
				stage 3 -  patch size                 &$16\times16$\\
				stage 3 -  hidden size               &768 \\
				stage 3 -  window size              &$7\times7$\\
				stage 3 -  number of heads     &24 \\
				\hdashline
				stage 4 -  patch size                 &$32\times32$\\
				stage 4 -  hidden size               &1536 \\
				stage 4 -  window size              &$7\times7$\\
				stage 4 -  number of heads     &48 \\			
				\hline			
				\multicolumn{2}{c}{\textsl{Text Encoder - BERT-Small}} \\
				\hline
				maximum length of word tokens  &60\\ 
				vocabulary size    							  &30522\\
				attention dropout probability       &0.1\\
				hidden activation function           &GELU\\
				hidden dropout probability    		&0.1\\
				initializer range                        	    &0.02\\
				intermediate size                     		&2048\\
				layer norm eps                         		   &$1e^{-12}$\\
				hidden size                                		 &512 \\
				number of attention heads                      	    &8 \\
				number of hidden layers        		&4 \\
				\hline
				
				\multicolumn{2}{c}{\textsl{Pre-Training}} \\
				\hline
				number of epochs &20 \\
				batch size &512 \\
				learning rate &2e-4 \\
				learning schedule &\verb|OneCycle| \\
				cycle momentum &Ture \\
				base momentum   &0.85\\
				max momentum   &0.95\\
				AdamW weight decay &0.01 \\
				AdamW $\beta_1$ &0.9 \\
				AdamW $\beta_2$ &0.999 \\
				\hline    
				
			\end{tabular}
		\vspace{-0.5cm}
	\end{table}

	\section{Hyperparameters and Implementation Details}\label{a3}
	In this section, we summarize the hyperparameters in our LOUPE model in Table~\ref{hyper}, including the hyperparameters of the image encoder, text encoder, and pre-training process. For the uncertainty-aware neural Shapley interaction learning module, we attempt three kinds of models~(\textsl{i.e., Conv1D, 3-Layer MLP + Attention, 3-Layer Transformer}) to implement it for token-level and semantics-level Shapley interaction approximation.
	
	For token-level Shapley interaction approximation, it takes the patch token sequence $\mathcal{X}^I = \{\mathbf{x}^I_i\}_{i=1}^{L_1}$, word token sequence $\mathcal{X}^T = \{\mathbf{x}^T_i\}_{i=1}^{L_2}$, and the visual region $\mathcal{R}_i = \{\mathbf{x}^I_{i, k} \}_{k=1}^{K_i}$ as input, and estimates the corresponding token-level Shapley interaction value for $\mathcal{R}_i$ along with the uncertainty $\sigma$.
	
	\textbf{Conv1D} model first performs \verb|Avg-Pooling| over learned patch representations of $\mathcal{R}_i$ to obtain the region representation $\mathbf{h}^I_i$, and then fuse the word and patch token representations with the region representation $\mathbf{h}^I_i$, respectively. Specifically, we project them into an unified semantic space by fully-connected layers and then fuse them through Hadamard product as:
	
	\begin{equation}
		\mathcal{F}^I = (\mathcal{W}_1 \mathbf{h}^I_i \mathfrak{1}^T ) \odot (\mathcal{W}_2 \mathcal{X}^I )
	\end{equation}
	
	where $\mathcal{W}_1$ and $\mathcal{W}_2$ are the learnable projection parameters, $\mathfrak{1}^T$ is the transpose of an all-ones vector, and $\odot$ represents Hadamard product. We can obtain $\mathcal{F}^T$ in a similar manner. Then, we apply 1D convolution operation with kernel size = 4 and stride = 2 over $\mathcal{F}^I$ and $\mathcal{F}^T$, respectively. Following with \verb|Max-Pooling| operation, we obtain $\tilde{\mathbf{f}}^I \in \mathbb{R}^d$ and $\tilde{\mathbf{f}}^T \in \mathbb{R}^d$. Next, we concatenate them with $\mathbf{h}^I_i$ and feed them to two separate 1-layer fully connected layers to get the Shapley interaction estimation and corresponding uncertainty.
	
	\textbf{3-Layer MLP + Attention} model first performs \verb|Avg-Pooling| over learned patch representations of $\mathcal{R}_i$ to obtain the region representation $\mathbf{h}^I_i$. Then, we use $\mathbf{h}^I_i$ as the query to attend the patch token sequence and compute a weighted sum of the patch token representations as:
	
	\begin{align}
		\tilde{\alpha}^I_j =& \mathcal{W}_3(tanh(\mathcal{W}_4 \mathbf{h}^I_i + \mathcal{W}_5 \mathbf{x}^I_{j}))\\
		\mathbf{\alpha}^I =& softmax([\tilde{\alpha}^I_1, ..., \tilde{\alpha}^I_{L_1}])\\
		\mathbf{e}^I =& \sum_{j=1} \alpha^I_i \mathbf{x}^I_j
	\end{align}
	
	Where $L_1$ is the number of patch tokens. We can obtain $\mathbf{e}^T$ for word token sequence in a similar manner. Consequently, we concatenate them and $\mathbf{h}^I_i$ and feed them to two separate 3-layer fully connected layers to get the Shapley interaction estimation and corresponding uncertainty.
	
	\textbf{3-Layer Transformer} model takes the concatenated sequence $\mathcal{X}^I$ and $\mathcal{X}^T$ as input. We add position embeddings and three kinds of token type embeddings~(\textsl{i.e., word token, context patch token, region patch token}) to them. We then apply three layers of transformer blocks to jointly encode the input sequence and take the output \verb|[CLS]| token to predict the Shapley interaction estimation and corresponding uncertainty, separately. 

	For semantics-level Shapley interaction approximation, it takes the $M$ regions $\mathcal{H}^I = \{\mathbf{h}^I_i\}_{i=1}^{M}$, $N$ phrases $\mathcal{H}^T = \{\mathbf{h}^T_j\}_{j=1}^{N}$, and the target region-phrase pair $<\mathbf{h}^I_i, \mathbf{h}^T_j>$ as input, and estimates the corresponding semantics-level Shapley interaction value for $<\mathbf{h}^I_i, \mathbf{h}^T_j>$ along with the uncertainty~$\sigma$. The architectures of the three models are consistent with their token-level implementations.

	\section{Pre-Training and Evaluation Details}\label{a4}
	
	\subsection{Pre-Training Dataset Details}
	
	As recent works~\cite{jia2021scaling, radford2021learning, yao2021filip} have shown that pre-training models can obtain great performance gain by scaling up the dataset, we construct a large-scale dataset, which consists of 240 million image-text pairs and covers a broad set of visual concepts. Concretely, we elaborate more details in the following.
	
	\noindent
	\textbf{Raw image-text pair collection.} We first harvest large-scale noisy image-text pairs from the web and design multiple filtering rules to improve the quality of the web data.
	
	\noindent
	\textbf{Image-based filtering.} Following ALIGN~\cite{jia2021scaling}, we remove pornographic images and keep only images where both dimensions are larger than 200 pixels. Also, we remove the images whose aspect ratio is larger than 10. To prevent from leaking testing data, we remove the images that appear in all downstream evaluation datasets (e.g., MSCOCO, Flickr30K). 
	
	\noindent
	\textbf{Text-based filtering.} We remove the repeated captions and keep only English texts. The texts that are shorter than 3 words or longer than 100 words are discarded. As ALIGN~\cite{jia2021scaling}, we also remove the texts that contain any rare token (outside of 100 million most frequent unigrams and bigrams from the raw dataset).
	
	\noindent
	\textbf{Joint image-text filtering.} Although the above filtering rules have filtered out many noisy data, it is hard to detect the mismatched image-text pairs, where the texts do not accurately describe the visual content of the images, resulting in undesirable noisy signals to vision-language pre-training. Inspired by BLIP~\cite{li2022blip}, we train a discriminator as a filtering model to predict whether the text is matched to the image. Specifically, the filtering model consists of an image encoder and an image-grounded text encoder, which takes the cross-attention to fuse image features and text features. The filtering model is trained on CC12M dataset using image-text contrastive loss and image-text matching loss.

	\subsection{Evaluation Details}
	\textbf{Zero-Shot Image-Text Retrieval.} We evaluate the zero-shot performance of LOUPE on the image-text retrieval task over the widely used Flickr30K~\cite{plummer2015flickr30k} and MSCOCO~\cite{lin2014microsoft} datasets. The image-text retrieval consists of two subtasks: image-to-text retrieval and text-to-image retrieval, where a model is required to identify an image from candidates given a caption describing its content, or vice versa. The MSCOCO dataset consists of 123,287 images, and each image is aligned with five captions. The Flickr30K dataset contains 31,783 images and five captions for each image. Following previous works~\cite{jia2021scaling, yao2021filip}, we evaluate the zero-shot performance on the 1K and 5K test sets of Flickr30K and MSCOCO, respectively. We take the final representation of \verb|[CLS]| tokens as the global representations of images and texts, and use them to measure the image-text similarity. We first compute the similarity scores for all image-text pairs. Then, we take the top-K candidates for ranking and report the top-K retrieval results.
	
	\textbf{Zero-Shot Transfer to Object Detection.} Without any fine-tuning, we evaluate the zero-shot transfer performance of LOUPE on the object detection task~\cite{ren2015faster} over the COCO~\cite{lin2014microsoft} and PASCAL VOC~\cite{everingham2008pascal} datasets. For the COCO Objects dataset, we use their 2017 validation split for evaluation. Previous zero-shot object detection models~\cite{bansal2018zero, rahman2020improved, zhu2020don} follow the split proposed by \cite{bansal2018zero}, which consists of 48 base classes and 17 novel classes. They train models on base classes and evaluate models on novel classes. Differently, we directly evaluate the zero-shot transfer performance on  both the base and novel classes, without fine-tuning on the base classes. Totally, we evaluate models on 4,836 test images that contain 33,152 instances of 65 classes. PASCAL VOC is a widely used object detection dataset, which contains 20 object classes. For PASCAL VOC, we evaluate models on 9657 instances of 5072 images. To complete object detection, we first use the region generation module to generate a set of candidate regions and then use prompt text (\textsl{i.e., an image of [object class name].}) to expand each detection label to a sentence. Next, we encode sentences for each object class by the learned text encoder and measure their similarity with the candidate regions as the classification scores. Following most zero-shot object detection methods, we use mean Average Precision~(mAP) at IoU of $\{0.3, 0.5\}$ as evaluation metrics.

	\textbf{Zero-Shot Transfer to Visual Grounding.} Visual grounding~\cite{yu2016modeling}~(also known as phrase localization and referring expression comprehension) aims to locate a specific visual region of the input image, according to the language referring expression. Visual grounding can be seen as generalized object detection, where the pre-defined class labels are replaced by language referring expression sentences. Without any fine-tuning, we evaluate the zero-shot transfer performance of LOUPE on the visual grounding task over the RefCOCO~\cite{yu2016modeling} and RefCOCO+~\cite{yu2016modeling} datasets. These two datasets are collected by the ReferitGame~\cite{kazemzadeh2014referitgame}, where a player is asked to write a language expression to refer to a specific object in the image, and another player is required to locate the target object given the image and the referring expression. RefCOCO dataset consists of 142,209 refer expressions for 50,000 objects in 19,994 images, which is split into train~(120,624 expressions), val~(10,834 expressions), testA~(5,657 expressions), testB~(5,095 expressions) sets. The images in testA set involve multiple persons and the images in testB set involve multiple objects. RefCOCO+ dataset consists of 141,564 expressions for 49,856 objects in 19,992 images, which is split into train~(120,191 expressions), val~(10,758 expressions), testA~(5,726 expressions), testB~(4,889 expressions) sets. We report the zero-shot transfer performance on the val, testA, and testB sets of both datasets.
	
	\begin{table}[!t]
		\caption{Image captioning evaluation results on COCO ``Karpathy'' test split.}
		\label{captioning}
		
		\centering
		\begin{tabular}{ l| cccc}
			\hline
			{\multirow{2}*{}}
			&\multicolumn{4}{c}{\textsl{Image Captioning}} 
			\\

			&BLEU@4 &METEOR &CIDEr &SPICE \\
			\hline 
			VLP~\cite{zhou2020unified}                                  &36.5            &28.4              &117.7                 &21.3  \\
			$\mathrm{OSCAR}_{\mathrm{large}}$~\cite{li2020oscar}             &37.4            &30.7              &127.8                 &23.5  \\
			$\mathrm{VinVL}_{\mathrm{large}}$~\cite{zhang2021vinvl}                 &38.5             &30.4             &130.8              &23.4 \\
			$\mathrm{BLIP}_{\mathrm{ViT-L}}$~\cite{li2022blip}                  &40.4             &-                    &136.7               &- \\
			$\mathrm{LEMON}_{\mathrm{large}}$~\cite{hu2021scaling}            &40.6             &30.4              &135.7               &23.5 \\
			\textbf{LOUPE}                                                        &\textbf{40.9}             &\textbf{31.5}                     &\textbf{137.8}                 &\textbf{24.3}      \\
			\hline
		\end{tabular}
		\vspace{-0.4cm}
	\end{table}
	
	\vspace{-0.3cm}
	\section{More Experiment Results on Vision-Language Generation Task}\label{a5}
	\vspace{-0.3cm}
	To further validate the generalization ability of the learned cross-modal representations by our LOUPE, we adapt the pre-trained LOUPE to vision-language generation task, \textsl{i.e.,} image captioning~\cite{anderson2018bottom}. Image captioning is the task of describing images with natural languages, which requires models to identify and describe the fine-grained semantics of images. The input images are encoded by the learned image encoder. As BLIP~\cite{li2022blip}, we train an image-grounded text decoder which shares the feed forward layers with the learned text encoder and adopts cross-attention to attend to the image features. The text decoder is trained with a language modeling loss to generate captions according to the images. 
	
	We evaluate the image captioning performance on the MSCOCO~\cite{lin2014microsoft} dataset, which is split into train~(113, 287 images), val~(5,000 images), ``Karpathy'' test split~(5,000 images). Each image has 5 captions. We use the train split to train the image-grounded text decoder and report the performance on the public ``Karpath'' 5k test split. Following standard metrics, we use BLEU@4, METEOR, CIDEr, and SPICE as evaluation metrics. We compare our LOUPE model with recent vision-language pre-training generation models~\cite{hu2021scaling, li2022blip, li2020oscar, zhang2021vinvl, zhou2020unified}. All methods are fine-tuned with cross-entropy loss only, without CIDEr optimization. As shown in Table~\ref{captioning}, our LOUPE achieves competitive performance on all metrics, which verifies the strong generalization ability of our model on downstream vision-language generation tasks.

	\begin{table}[!t]
		\caption{Further ablation results (R@1) with respect to different image encoders.}
		\label{t_image_encoder}
		\resizebox{\textwidth}{!}{
			\centering
			\begin{tabular}{ ll| c| cc |cc}
				\hline
				\multicolumn{2}{l|}{\multirow{2}*{}} 
				&\multirow{2}*{Image Encoder}
				&\multicolumn{2}{c|}{\textsl{Flickr30K}} 
				&\multicolumn{2}{c}{\textsl{MSCOCO}}
				\\
				
				\multicolumn{2}{l|}{} & &\textsl{image-to-text}&\textsl{text-to-image} &\textsl{image-to-text} &\textsl{text-to-image}\\
				\hline
				
				\multicolumn{2}{l|}{1\quad$\mathrm{ALIGN}$~\cite{jia2021scaling}} &EfficientNet &88.6    &75.7   &58.6   &45.6  \\
				\multicolumn{2}{l|}{2\quad$\mathrm{FILIP}$~\cite{yao2021filip}}  &ViT-L  &89.8   &75.0   &61.3  &45.9   \\
				\multicolumn{2}{l|}{3\quad$\mathrm{CLIP}$~\cite{radford2021learning}}  &ViT-L &88.0  &68.7  &58.4 &37.8 \\
				\multicolumn{2}{l|}{4\quad${\mathrm{CLIP}}^*$}  &Swin-L  &88.7  &74.3  &59.3  &46.2  \\
				\multicolumn{2}{l|}{5\quad\textbf{LOUPE}} &Swin-L  &\textbf{90.5}    &\textbf{76.3}   &\textbf{62.3}   &\textbf{50.1}   \\
				\hline    
				
			\end{tabular}
		}
		\vspace{-0.4cm}
	\end{table}

	\vspace{-0.3cm}
	\section{Further Analysis on the Image Encoder}\label{a6}
	\vspace{-0.3cm}
	
	In our work, we adopt the Swin-L~\cite{liu2021Swin} as our image encoder due to the following considerations. (1)~The shifted windowing scheme of Swin Transformer achieves linear computational complexity with respect to image size, which is more efficient than ViT~\cite{dosovitskiy2020image}. This merit is particularly beneficial to the vision-language pre-training as we need to process  large-scale images (240M). (2)~The hierarchical architecture of Swin Transformer is more flexible to model semantic regions at various scales. 
	
	To further verify the performance gain from our proposed fine-grained semantically aligned vision-language pre-training framework, we implement a variant version of $\mathrm{CLIP}$ that adopts Swin-L as the image encoder~(Row 4 in Table~\ref{t_image_encoder}), using the same training dataset as our LOUPE. It can also be viewed as the backbone of our LUOPE~(without optimization from our token-level and semantics-level Shapley interaction modeling). As shown in Table~\ref{t_image_encoder}, comparing $\mathrm{CLIP}^*$ with $\mathrm{CLIP}$, the Swin-L image encoder does bring some improvements over $\mathrm{CLIP}$. However, there is still a clear performance gap between $\mathrm{CLIP}^*$ and our LOUPE. With the same architecture, our LOUPE has 2.68 points higher average R@1 than the $\mathrm{CLIP}^*$ over two datasets. This further verifies that the main performance gain comes from our proposed fine-grained semantically aligned vision-language pre-training framework. Notably, we observe that the text-to-image retrieval of our implementation is obviously higher than $\mathrm{CLIP}$. This phenomenon has also been confirmed by~\cite{jia2021scaling, yao2021filip} (see Row 1 and Row 2 in Table~\ref{t_image_encoder}). We suppose that it might be caused by some training details or the dataset collection of $\mathrm{CLIP}$.
	
	\vspace{-0.3cm}
	\section{More Qualitative Examples on Object Detection and Visual Grounding}\label{a7}
\begin{figure}[!t]
		\centering
		\includegraphics[width=\textwidth]{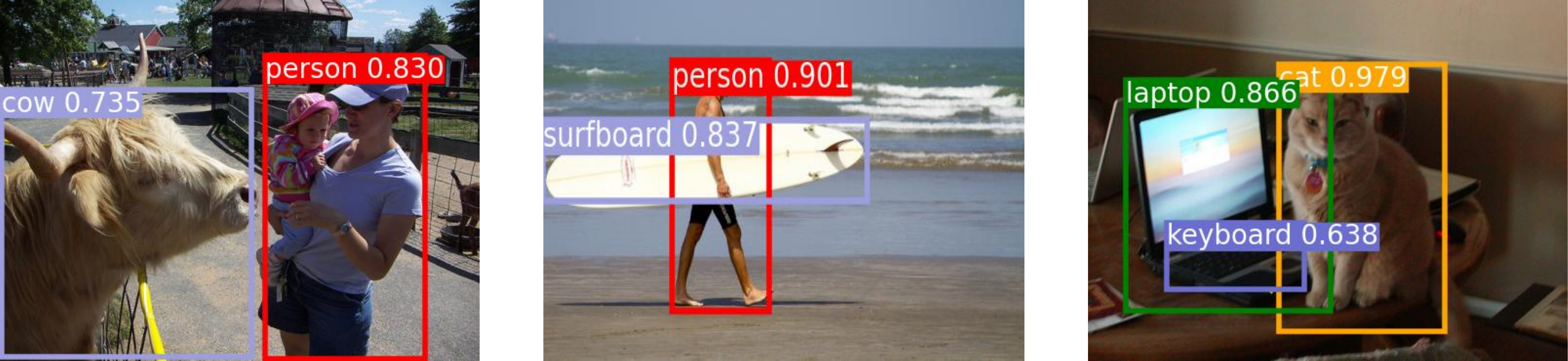}
		\caption{Qualitative examples of object detection on COCO Objects dataset.}
		\label{qualitative1}
	\end{figure}

\begin{figure}[!t]
		\centering
		\includegraphics[width=\textwidth]{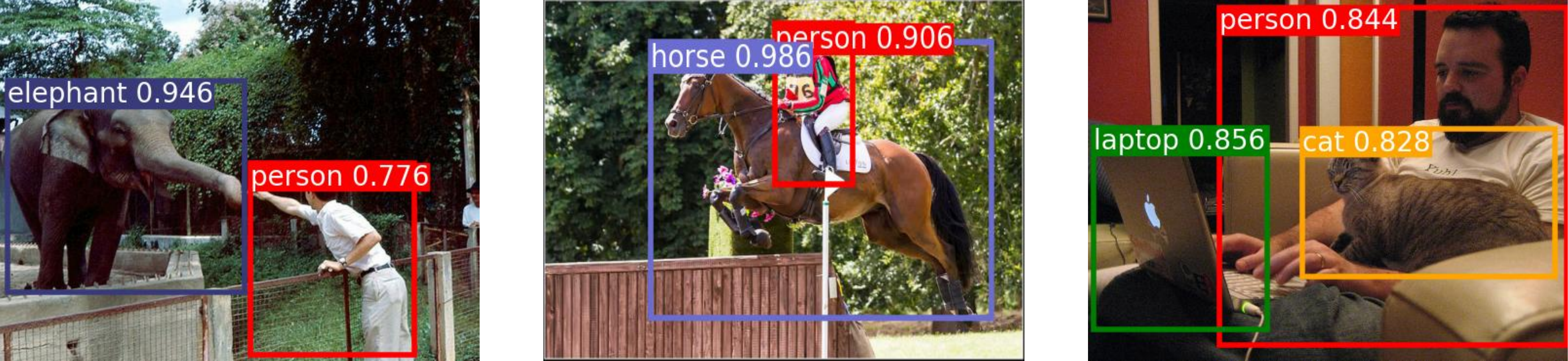}
		\caption{Qualitative examples of object detection on PASCAL VOC dataset.}
		\label{qualitative2}
	\end{figure}

\begin{figure}[!t]
		\centering
		\includegraphics[width=\textwidth]{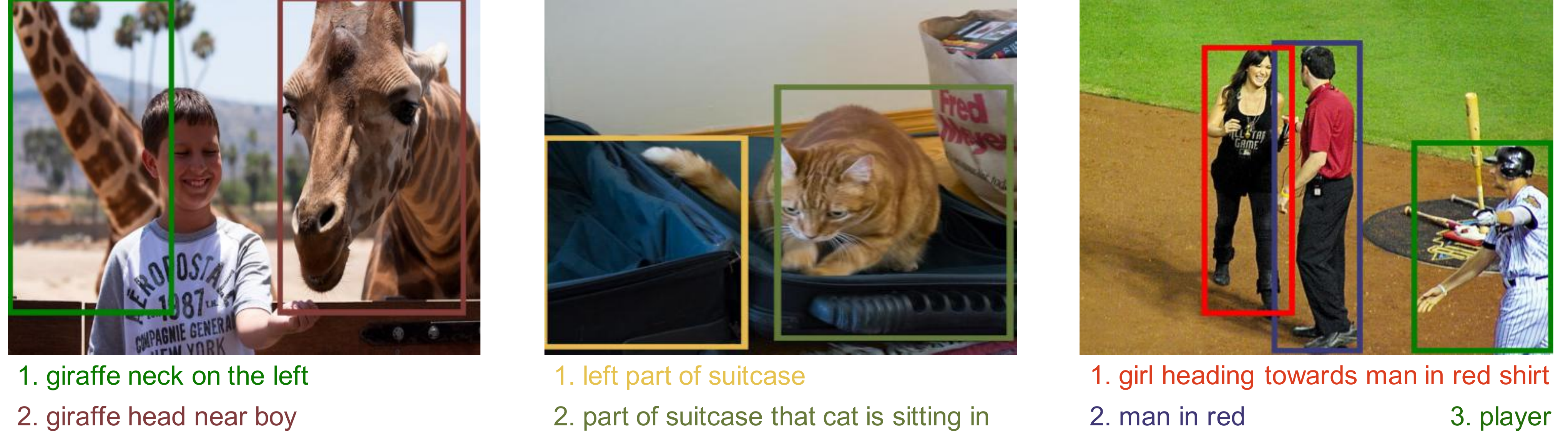}
		\caption{Qualitative examples of visual grounding on RefCOCO dataset.}
		\label{qualitative3}
	\end{figure}

\begin{figure}[!t]
		\centering
		\includegraphics[width=\textwidth]{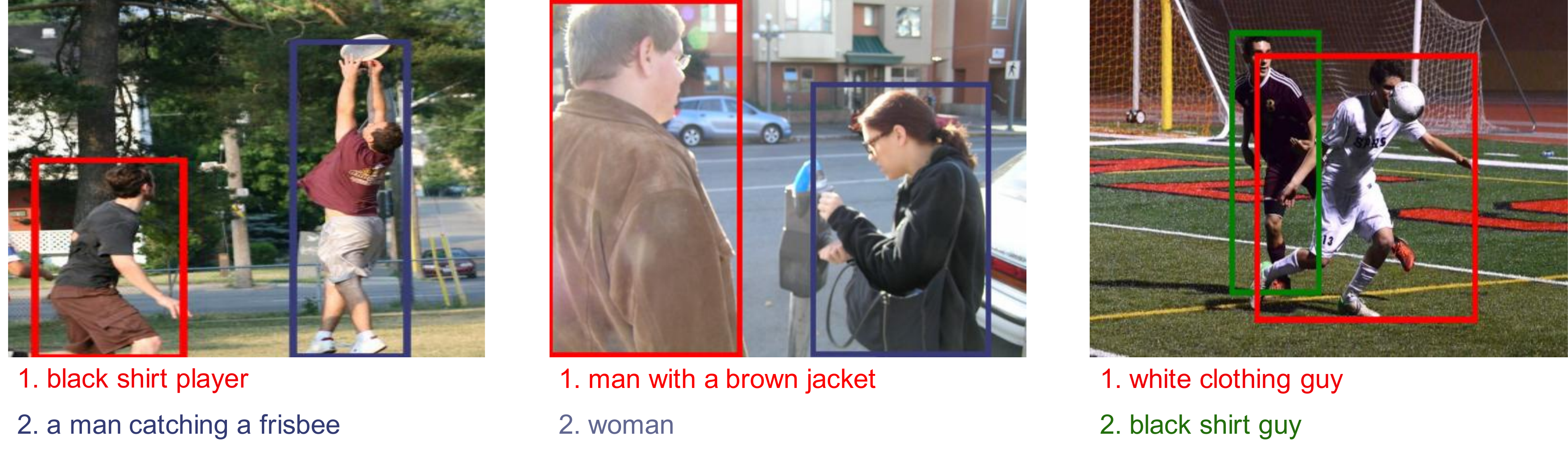}
		\caption{Qualitative examples of visual grounding on RefCOCO+ dataset.}
		\label{qualitative4}
	\end{figure}

	\vspace{-0.3cm}
	For a more intuitive view of the performance of our model on object detection and visual grounding, we visualize more qualitative examples. Concretely, Figure~\ref{qualitative1} and Figure~\ref{qualitative2} show more object detection examples on the COCO~\cite{lin2014microsoft} and PASCAL VOC~\cite{everingham2008pascal} datasets. Figure~\ref{qualitative3} and Figure~\ref{qualitative4} show more visual grounding examples on the RefCOCO~\cite{yu2016modeling} and RefCOCO+~\cite{yu2016modeling} datasets.
	
	\section{Linear Probing Evaluation}\label{a9}
	In this section, we evaluate the linear probing performance of our LOUPE on image classification. Following the same evaluation setting as CLIP~\cite{radford2021learning}, we freeze the whole backbone network and only fine-tuning the last linear classification layer, which takes the \verb|[CLS]| token as input. We report the linear probing performance over 11 datasets in Table~\ref{linear_prob}. Our LOUPE outperforms CLIP with average improvement of 1.6\%. Notably, on ImageNet, the largest dataset among 11 datasets, our LOUPE surpasses CLIP by 1.8\%.
	\begin{table}[!t]
		\caption{Linear probing performance (top-1 accuracy) over 11 datasets.}
		\label{linear_prob}
			\centering
			\begin{tabular}{ l| ccccccccccc}
				\hline
				&\rotatebox{90}{CIFAR10}  &\rotatebox{90}{Food101} &\rotatebox{90}{StanfordCars} &\rotatebox{90}{SUN397} &\rotatebox{90}{Flowers102} &\rotatebox{90}{Country211} &\rotatebox{90}{FER2013} &\rotatebox{90}{Aircrafts} &\rotatebox{90}{OxfordPets} &\rotatebox{90}{Caltech101} &\rotatebox{90}{ImageNet}
				\\

				\hline
				
				CLIP     &\textbf{98.0}   &95.2   &90.9    &81.8   &99.2   &46.4 &\textbf{72.9}   &69.4   &95.1 &96.5   &83.9  \\
				\textbf{LOUPE}     &97.6  &\textbf{96.0}   &\textbf{92.1}    &\textbf{82.6}   &\textbf{99.5}   &\textbf{49.3}  &70.7   &\textbf{80.2}  &\textbf{95.5}  &\textbf{97.5}   &\textbf{85.7} \\
				
				\hline    
				
			\end{tabular}
			
	\end{table}

	\begin{table}[!t]
		\caption{Comparison of training cost  and architecture parameters.}
		\label{cost}
		\resizebox{\textwidth}{!}{
			\centering
			\begin{tabular}{ ll | c | c| ccc}
				\hline
				\multicolumn{2}{l|}{} 
				&Pre-Training Image-Text Pairs
				&Parameters
				&GPUs
				&Days
				&GPU Days
				
				\\
				
				\hline
				
				\multicolumn{2}{l|}{CLIP}       &400M     &425M      &256 V100    &12 days    &3072   \\
				\multicolumn{2}{l|}{ALIGN}    &1800M  &820M   &1024 TPUv3    &-       &- \\
				\multicolumn{2}{l|}{FILIP}     &340M    &417M   &192 V100    &24 days    &4608   \\
				
				\multicolumn{2}{l|}{\textbf{LOUPE}}  &240M &226M    &128 V100   &20 days     &2560  \\
				
				\hline    
				
			\end{tabular}
			
		}
	\end{table}

	\section{Training Efficiency Discussion}\label{a10}
	Although our proposed Shapley interaction modeling increases the training time per iteration, it enables our model to converge with fewer total iterations by encouraging our model to learn fine-grained region-phrase alignment beyond coarse image-text alignment. As shown in Table~\ref{cost}, our LOUPE achieves the best performance while using relatively small GPU days (128 GPUs $\times$ 20 days).
	
	Indeed, the proposed Shapley interaction modeling increases the training time per iteration, but it enables our model to learn fine-grained region-phrase alignment from raw image-text pairs without any object-level human annotations. Our LOUPE can be used as a zero-shot object detector without any fine-tuning. Compared with the expensive cost of human annotations, the increased training time might be acceptable. Meanwhile, manual annotations for extremely diverse object categories in the real world are unscalable and even impossible while our model demonstrates a promising alternative, that is, learning fine-grained semantics from raw texts about images, which are easily available and contain a broader set of visual concepts. For example, the right case of Figure 4 in the main paper shows that LOUPE successfully recognizes the leash region and aligns it with the “a leash” phrase. Note that the “leash” category has never appeared in any existing object detection datasets.
	
	On the other hand, our method is much more efficient than methods that rely on off-the-shelf object detectors (e.g., Faster R-CNN) to extract visual features. Recent studies~\cite{kim2021vilt, yao2021filip} have noticed that extracting visual features using object detectors greatly slows down the training (about 20 FPS per GPU) and requires more GPU memory. Thus, our model avoids such a heavy burden while being able to identify semantic-rich visual regions without any pre-training detectors or human annotations.

	\begin{figure}[!h]
		\centering
		\includegraphics[width=\textwidth]{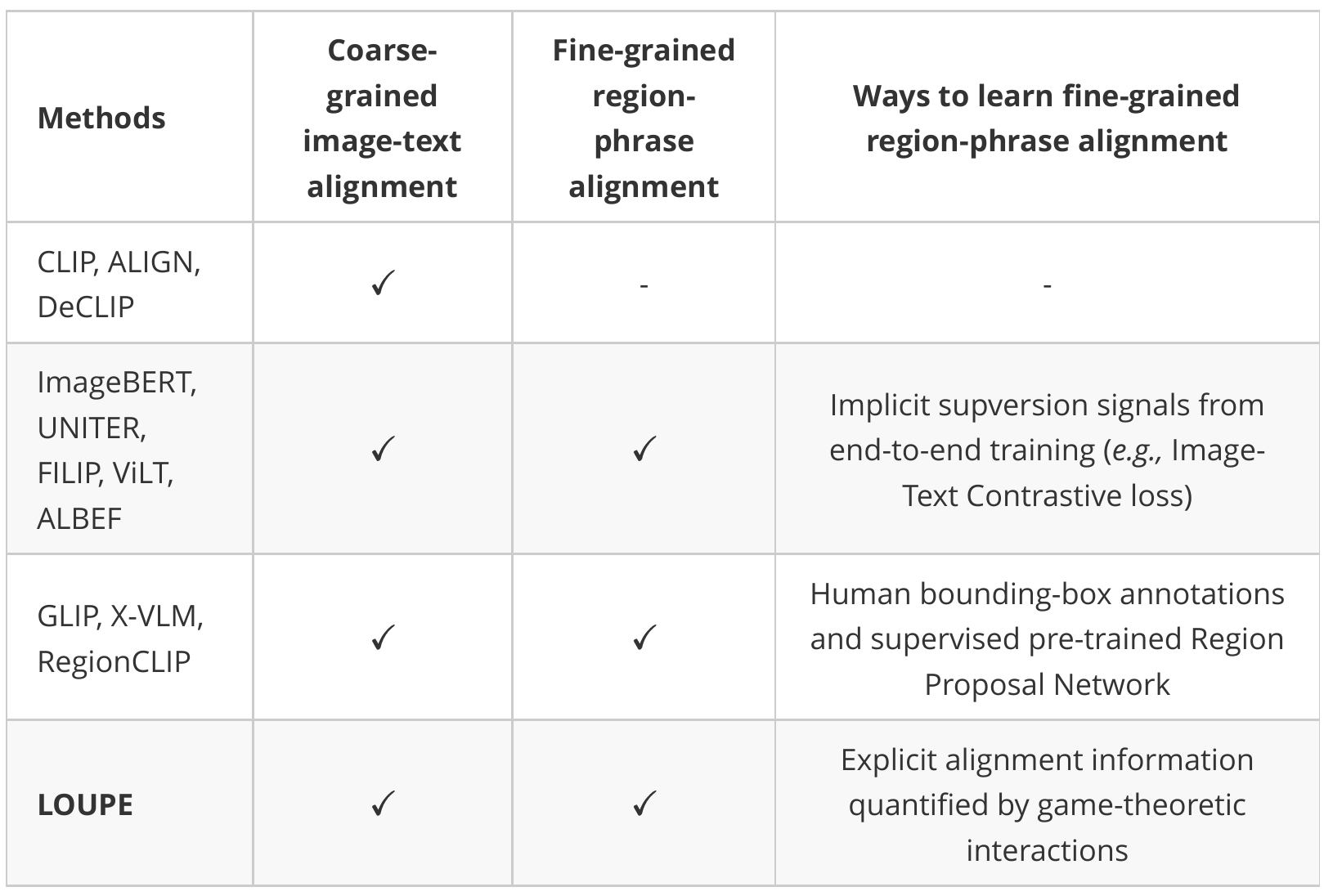}
		\caption{Comparison of the LOUPE with existing methods.}
		\label{comparison}
	\end{figure}

	\section{Detailed Discussion with Some Related Works}\label{a11}
	In this section, we first provide comparison table to highlight key differences of our LOUPE with various methods. Then, we provide a detailed discussion with three recent works~(\textsl{i.e.}, FILIP~\cite{ yao2021filip}, RegionCLIP~\cite{zhong2022regionclip}, X-VLM~\cite{zeng2021multi}), which also investigate fine-grained semantic alignment. 
	
	We highlight key differences in Figure~\ref{comparison}. Our LOUPE differs as it explicitly learns fine-grained region-phrase alignment from the novel perspective of game-theoretic interactions, without resorting to any object-level human annotations and pre-trained Region Proposal Network (RPN). Notably, the human bounding-box annotations are usually limited to the pre-defined object categories, and the RPN can only detect regions belonging to the pre-defined categories of pre-training object detection datasets. Thus, the methods that use human bounding-box annotations or pre-trained RPN usually suffer from detecting novel objects beyond the pre-defined categories while LOUPE learns from large-scale raw image-text pairs, which are more scalable and contain a broader set of visual concepts.
	
	Compared with FILIP, the superiorities of using Shapley Interaction modeling are mainly three-fold: \textbf{1)}~We suppose that directly computing token-wise alignment between every patch token and word token is not efficient and meaningful because an individual word token or patch token might not contain complete semantics. A semantic-rich phrase (e.g., “a girl in a blue coat”) usually consists of multiple words, and its corresponding visual region is composed of multiple patches. Also, some words (e.g., "is", "the") and patches (e.g., background pixels) are not meaningful. Based on this insight, our LOUPE differs as we first propose token-level Shapley interaction modeling to aggregate patches into semantic-meaningful regions, and then introduce semantics-level Shapley interaction modeling to explicitly model the fine-grained semantic alignment between semantic-meaningful regions and phrases. \textbf{2)}~Although FILIP computes token-wise similarity to simulate the fine-grained alignment, it can only learn implicit alignment from the supervision of image-text contrastive loss, lacking training signals to explicitly encourage semantic alignment between visual regions and textual phrases. In contrast, our Shapley interaction modeling provides explicit supervision signals (e.g., the alignment matrices visualized in Figure 4) to learn the fine-grained alignment. The consistently superior performance of our LOUPE than FILIP over all metrics also demonstrates the benefit of explicit fine-grained alignment learning. \textbf{3)} FILIP can not be directly applied to object detection and visual grounding through implicit token-wise alignment learning while our LOUPE can immediately transfer to these tasks without any fine-tuning. It is because the proposed Shapley interaction modeling enables our model to identify semantic regions and align these regions with language. As shown in Table 2, without any bounding-box annotations and fine-tuning, our LOUPE achieves competitive performance across four object detection and visual grounding benchmarks.
	
	Our LOUPE is different from RegionCLIP in the following aspects: \textbf{1)}~RegionCLIP uses pre-trained Region Proposal Network (RPN) to detect regions in images. However, RPN is usually pre-trained on pre-defined object categories (e.g., 80 classes for MSCOCO), which can not cover extensive categories of objects in the large-scale pre-training dataset. Furthermore, since the RPN casts excessive demand on memory and computation, existing methods (i.e., RegionCLIP) usually fix the parameters of RPN and regard region detection as pre-processing step, disconnected with vision-language pre-training. Thus, the performance of RegionCLIP is also restricted by the quality of the RPN. In contrast, our LOUPE learns to identify semantic regions of images by token-level Shapley interaction modeling, which is more scalable and enables our LOUPE to learn a broader set of visual concepts from large-scale pre-training dataset. \textbf{2)}~RegionCLIP constructs a pool of object concepts from image-text corpus and aligns visual regions with these concepts. These concepts are usually individual nouns (e.g., boy, kite, bus). In contrast, our LOUPE focuses on phrases that involve rich context (e.g., "a boy running on the grass"). By aligning visual regions with phrases that contain rich semantic context, our LOUPE can learn a boarder set of visual concepts (e.g., objects, actions, relations) from the large-scale pre-training dataset.
	
	As for X-VLM, the main differences lie in three-fold: \textbf{1)}~X-VLM is trained on well-annotated datasets, where regions with bounding-box annotations are provided and each of them is associated with a description text. Such a manner is time-consuming and hard to scale to larger raw image-text data from the web. Our LOUPE differs as we are trained on noisy image-text pairs from the Internet. \textbf{2)}~X-VLM takes ground-truth regions as input and is trained to predict the bounding-box supervised by the regression loss on the ground-truth coordinates. In contrast, our LOUPE learns to identify semantic regions of images without such strong supervision signals from human annotations. \textbf{3)}~X-VLM has ground-truth alignment information between regions and their corresponding description texts, which provide strong supervision signals for region-text matching. By comparison, our LOUPE learns the fine-grained region-phrase alignment from game-theoretic interactions.

	\clearpage
	
	{
		\small
		\bibliographystyle{ieee_fullname}
		\bibliography{neurips}
	}
	
	\clearpage

\end{document}